\documentclass[10pt,journal,compsoc]{IEEEtran}

\ifCLASSOPTIONcompsoc
  \usepackage[nocompress]{cite}
\else
  \usepackage{cite}
\fi

\ifCLASSINFOpdf
\else
\fi

\usepackage{url}            
\usepackage{booktabs}       
\usepackage{amsfonts}       
\usepackage{nicefrac}       
\usepackage{microtype}      
\usepackage{xcolor}         
\usepackage{graphicx}
\usepackage{amsmath,amssymb}
\usepackage{verbatim}
\usepackage{multirow}
\usepackage{marvosym}
\usepackage{makecell}
\usepackage{subcaption}
\usepackage{bm}
\usepackage{wrapfig}
\usepackage{caption}
\usepackage{algorithm}
\usepackage{algorithmic}
\usepackage[marginal]{footmisc}

\begin{document}
%
\title{Pragmatic Communication in Multi-Agent Collaborative Perception}

\author{Yue Hu, Xianghe Pang, Xiaoqi Qin, Yonina C. Eldar, Siheng Chen, Ping Zhang, Wenjun Zhang
\IEEEcompsocitemizethanks{\IEEEcompsocthanksitem Y. Hu, X. Pang, S. Chen, W, Zhang are with the Cooperative Medianet Innovation Center (CMIC), Shanghai Jiao Tong University, Shanghai, China. E-mail: 18671129361, xianghep, sihengc, zhangwenjun@sjtu.edu.cn. 

\IEEEcompsocthanksitem X. Qin and P. Zhang are with the School of Information and Communication Engineering, Beijing University of Posts and Telecommunication, E-mail: xiaoqiqin, pzhang@bupt.edu.cn.

\IEEEcompsocthanksitem Y. C. Eldar is with the Department of Mathematics and Computer Science, Weizmann Institute of Science, Rehovot, Israel, E-mail: yonina.eldar@weizmann.ac.il.

\IEEEcompsocthanksitem Corresponding author is Siheng Chen. \protect \\
}
}


\IEEEtitleabstractindextext{%
\begin{abstract}

Collaborative perception allows each agent to enhance its perceptual abilities by exchanging messages with others. It inherently results in a trade-off between perception ability and communication costs. Previous works transmit complete full-frame high-dimensional feature maps among agents, resulting in substantial communication costs. To promote communication efficiency, we propose only transmitting the information needed for the collaborator's downstream task. 
This pragmatic communication strategy focuses on three key aspects: 
i) pragmatic message selection, which selects task-critical parts from the complete data, resulting in spatially and temporally sparse feature vectors; 
ii) pragmatic message representation, which achieves pragmatic approximation of high-dimensional feature vectors with a task-adaptive dictionary, enabling communicating with integer indices;
iii) pragmatic collaborator selection, which identifies beneficial collaborators, pruning unnecessary communication links.
Following this strategy, we first formulate a mathematical optimization framework for the perception-communication trade-off and then propose~\texttt{PragComm}, a multi-agent collaborative perception system with two key components: i) single-agent detection and tracking and ii) pragmatic collaboration. The proposed~\texttt{PragComm} promotes pragmatic communication and adapts to a wide range of communication conditions.
We evaluate~\texttt{PragComm} for both collaborative 3D object detection and tracking tasks in both real-world, V2V4Real, and simulation datasets, OPV2V and V2X-SIM2.0.~\texttt{PragComm} consistently outperforms previous methods with more than $32.7$K$\times$ lower communication volume on OPV2V. Code is available at~\url{github.com/PhyllisH/PragComm}.

\end{abstract}


}

\maketitle

\IEEEdisplaynontitleabstractindextext

\IEEEpeerreviewmaketitle


\begin{figure*}[!t]
    \centering
    \includegraphics[width=1.0\linewidth]{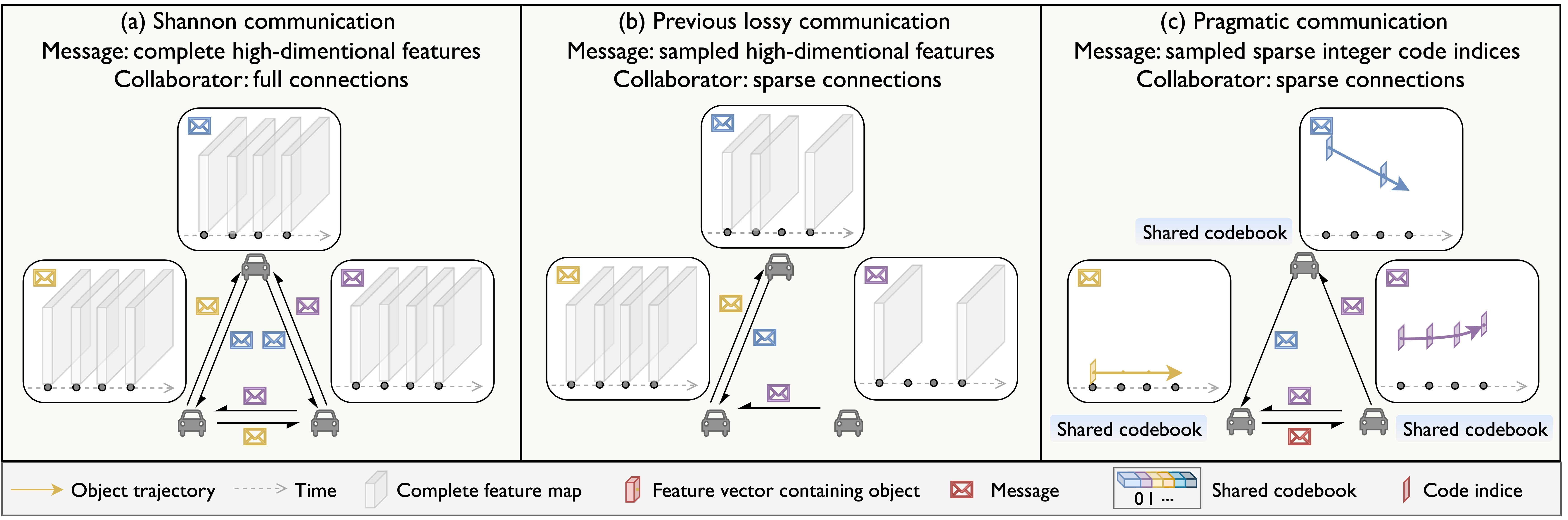}
    \vspace{-6mm}
    \caption{Communication strategies in collaborative perception. 
    (a) Shannon communication employs lossless
compression techniques to compress data to messages and is lossless for general tasks while causing substantial communication costs. (b) Previous communication methods leverage lossy compression and follow an all-or-nothing strategy to compress task-critical and task-irrelevant data without distinction, compromising task utility. (c) Our pragmatic communication retains only task-critical data with code indices in the pragmatic messages, this is, transmitting the demanded foreground object dynamics to each collaborator, reducing communication costs while retaining task utility.
    }
    \label{fig:introduction}
    \vspace{-6mm}
\end{figure*}

\IEEEraisesectionheading{\section{Introduction}\label{sec:introduction}}

\IEEEPARstart{M}ulti-agent collaborative perception targets to achieve more holistic perception by enabling agents to exchange complementary perceptual information through communication. 
Collaborative perception ensures expansive visibility, seeing through obstacles and identifying small, distant targets, thus achieving a thorough understanding of the environment. It provides a promising direction to fundamentally overcome the physical limitations of single-agent perception, such as limited field of view, occlusion, and long-range issues. As the forefront of autonomous systems, collaborative perception can enhance perception capabilities and further improve system-wide functionality and safety across various real-world applications, including autonomous driving~\cite{v2vnet,disconet,Chen20213DPC}, robotics~\cite{li2020mechanism,zaccaria2021multi}, and unmanned aerial vehicles (UAVs) for search and rescue missions~\cite{scherer2015autonomous,alotaibi2019lsar,DVDET}. 

Here, we mainly focus on the collaborative perception task of detection and tracking where each agent has its specific objective to identify the foreground object sequences over time within its designated perception region. 
To achieve its task, each agent serves dual roles: as a supporter, using communication resources to provide additional information through messages, and as a receiver, improving perception capability by utilizing these messages.
In this task, a central challenge is optimizing the trade-off between perception performance and communication cost~\cite{LiuWhen2com:CVPR20,LiuWho2com:ICRA20,HuWhere2comm:NeurIPS22,HuCollaboration:CVPR23,WangV2vnet:ECCV20,LiLearning:NeurIPS21}. 
The communication cost increases linearly with the size of perceptual regions and the duration of time, and quadratically with the number of collaborating agents. 
Despite advancements in recent 6G communication systems~\cite{Alwis6G:JCS2021}, practical communication resources are always constrained, which makes it difficult to support the extensive gigabyte-sized communication required for continuously sharing all perceptual data in real-time among all agents. 
As a result, the application of collaborative perception is limited to a small number of agents and a limited timeframe, thereby resulting in only marginal performance improvements compared to single-agent perception.

To address this challenge, the key lies in optimizing messages to fill each agent's specific perception task demand within the communication budget. 
A straightforward way is using source coding in the traditional Shannon communication paradigm~\cite{Shannon1948AMT,MacKay2004InformationTI,Wyner1976TheRF,Cover2005ElementsOI,Huffman1952AMF,Leeuwen1976OnTC,Salomon2004DataCT}.
This approach encodes original data into a sequence of codes, assigning shorter codes to frequent data and longer codes to rare ones, creating a compact representation without information loss. 
In the context of collaborative perception, this approach efficiently compresses perception data into messages at the supporter's end and ensures lossless reproduction at the receiver's end. 
This approach enhances communication efficiency while preserving the utility for general downstream tasks, including perception. 
However, this Shannon paradigm has fundamental limitations in scenarios requiring communication tailored to specific downstream tasks, as it inevitably wastes resources on irrelevant data. 
For instance, in collaborative vehicle detection tasks with camera-sensor-equipped agents, the Shannon paradigm encodes each pixel uniformly, not differentiating between non-essential background and crucial vehicle pixels. These background pixels hugely waste communication resources without aiding detection performance, thus impacting the perception-communication trade-off.

To overcome this sub-optimality, we propose a novel pragmatic communication strategy. 
Here, communication refers to the process where data is compressed into messages on the supporter side, and the messages are reconstructed to data on the receiver side, without the interference of the noisy channel in the traditional Shannon paradigm.
Pragmatic reveals that each part of the data is correlated to the particular utility of the specific downstream task.
This strategy's core idea is to acknowledge the pragmatic significance of the data, where the Shannon paradigm initially overlooks, and to develop pragmatic messages that retain only the data necessary for the collaborator's downstream task.
Specifically, in the collaborative detection and tracking task, pragmatic messages are crafted from three aspects:
First, message selection focuses on selecting specific parts demanded by the collaborator's task from the complete data. For instance, the sparse areas containing the collaborator's invisible foreground objects and key timestamps that highlight object dynamics should be prioritized. 
Second, message representation focuses on forming compact representations by compressing task-irrelevant information more heavily and task-relevant information slightly. 
For instance, an object's detailed appearance might be compressed more than its boundaries, which are crucial for detection.
Third, message transmission (collaborator selection) focuses on transmitting these messages to agents whose demanded information is present within them. 
This strategy selectively discards a large portion of task-irrelevant data to craft pragmatic messages. By doing so, the communication volume amount breaks Shannon's limits, leading to enhanced communication efficiency while maintaining perception utility.

\begin{figure*}[!t]
    \centering
    \includegraphics[width=1.0\linewidth]{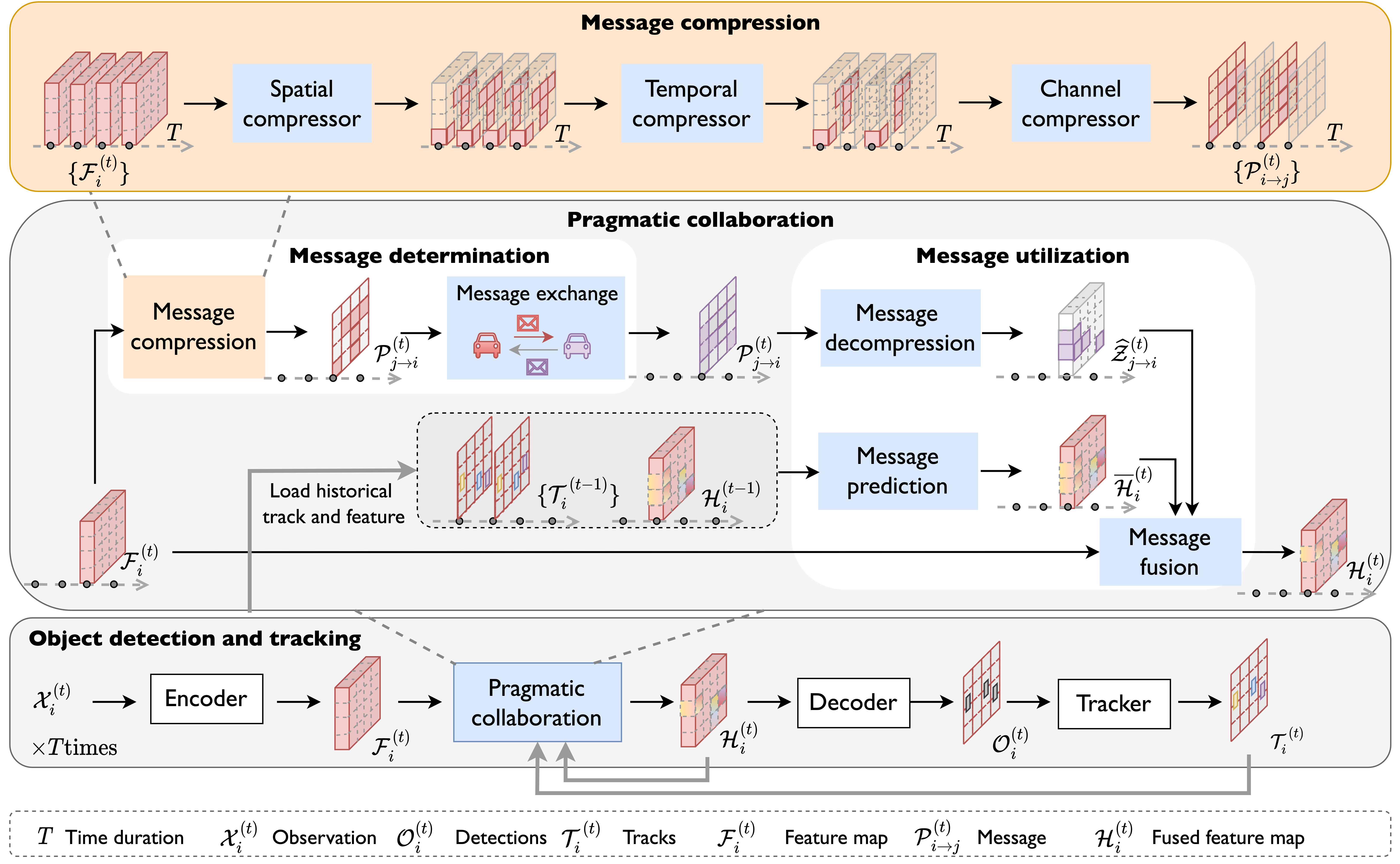}
    \vspace{-6mm}
    \caption{PragComm achieves a collaborative object detection and tracking system. Collaboration enhances individual perceptual features with communication-enabled pragmatic messages, and message compression ensures efficient communication.}
    \label{fig:framework}
\vspace{-6mm}
\end{figure*}

Following this strategy, we propose \texttt{PragComm}, a novel pragmatic collaborative perception system, which leverages pragmatic messages to enhance detection and tracking capabilities of multiple collaborative agents; see Fig.~\ref{fig:framework}.
This system includes two key components: 1) single-agent detection and tracking, which extracts high-dimensional perceptual features from sensor inputs and decodes them into perceptual outputs, offering basic detection and tracking capabilities, and 2) pragmatic collaboration, where each agent performs dual roles: as a supporter, compress high-dimensional perceptual features into pragmatic messages, and as a receiver, reconstruct perceptual features from these messages and integrate them to enhance collaborative perception capabilities.
Specifically, the pragmatic messages are determined and utilized with three key steps.
First, pragmatic message selection encompasses spatial and temporal dimensions, selecting key foreground regions and significant timestamps. Spatially, the spatial compressor on the supporter side employs a confidence map to assess and select sparse areas containing objects. Temporally, on the supporter and receiver sides, the temporal compressor and the message predictor collaboratively sample and recover object sequences by modeling object movements. The temporal compressor disregards timestamps without movement changes, while the predictor, even in the absence of received messages at these times, reconstructs the object's state using the established movement model.
Second, pragmatic message representation focuses on channel dimension, transforming the high-dimensional features to integer code indices. A task-adaptive codebook that preserves essential perceptual features is constructed and shared among agents. On the supporter side, the channel compressor approximates the perceptual feature vector using the nearest codes from the codebook, whereas, on the receiver side, the message decompression module reconstructs the feature by querying the shared codebook given the received code-index-based messages.
Third, pragmatic collaborator selection focuses on the communication connection aspect, establishing only necessary links. Each agent's needs are deduced from its past messages. The message exchange module forms communication links only when the supporter can offer the data the receiver requires, ensuring mutually beneficial communication.

\texttt{PragComm} offers five distinct advantages.
First, \texttt{PragComm} promotes pragmatic communication while previous works~\cite{XuOPV2V:ICRA22,XuV2XViT:ECCV22,WangV2vnet:ECCV20,LuRobust:ICRA23,YangSpatioTemporalDA:ICCV2023} follow the traditional Shannon communication paradigm and employ lossless compression techniques, exemplified by V2VNet~\cite{WangV2vnet:ECCV20} with source-coding. \texttt{PragComm} discards irrelevant data, thereby improving communication efficiency. 
Second, \texttt{PragComm} considers a comprehensive pragmatic communication while previous works~~\cite{LiLearning:NeurIPS21,LiuWhen2com:CVPR20,LiuWho2com:ICRA20} that apply lossy compression techniques only consider partial aspects. 
DiscoNet~\cite{LiLearning:NeurIPS21} leverages 1D convolution for dimension reduction, and When2com~\cite{LiuWhen2com:CVPR20} and Who2com~\cite{LiuWho2com:ICRA20} explores collaborator selection with attention mechanisms.
While they lack the consideration of the pragmatic significance of the data, compressing both critical and irrelevant data without distinction and thus compromising perception utility, \texttt{PragComm} emphasizes more compression on irrelevant data and less on crucial information. This targeted task-specific compression enhances communication efficiency while preserving perception utility.
Third, \texttt{PragComm} considers the communication over the entire temporal duration while all previous works~\cite{XuOPV2V:ICRA22,XuV2XViT:ECCV22,LiLearning:NeurIPS21,LiuWhen2com:CVPR20,LiuWho2com:ICRA20,HuWhere2comm:NeurIPS22,HuCollaboration:CVPR23,YangWhat2comm:ICCV2023,WangUMCAU:CVPR2023,How2comm:NeurIPS2023} separately consider communication at each timestamp. This comprehensive view takes advantage of the inherent temporal coherence of the targeted object sequences. It not only minimizes temporal redundancy, improving communication efficiency, but also leverages complementary messages across different times, promoting perception performance.
Fourth, \texttt{PragComm} adapts to the entire range of communication conditions by flexibly selecting spatial regions, timestamps, and code index compositions for representation, while previous works~\cite{XuOPV2V:ICRA22,XuV2XViT:ECCV22,LiLearning:NeurIPS21,LiuWhen2com:CVPR20,LiuWho2com:ICRA20,HuWhere2comm:NeurIPS22,HuCollaboration:CVPR23} only handle one predefined communication bandwidth.
Fifth, \texttt{PragComm} uses deep neural networks for optimizing task-based messages, while previous works~\cite{ZhangHardwareIO:TIE2023,XiBiLiMOBM:TSP2020,BernardoDesignAA:TSP2022} focus on task-based sampling and quantization within the traditional framework of multiple-input multiple-output (MIMO) communication channels. \texttt{PragComm}, differing in direction and technique, complements these studies and offers the potential to elevate communication efficiency to a higher level.





To validate PragComm, we conduct extensive experiments on collaborative 3D object detection and tracking tasks. Our evaluation encompasses four datasets, including both real-world and simulation datasets, multiple modalities such as camera and LiDAR, and diverse agent types, including vehicles and infrastructures, to ensure the versatility of our approach.
Experiment results show that: i) PragComm substantially improves communication efficiency while maintaining high-performance levels by leveraging the pragmatic message selection and representation; ii) PragComm consistently achieves superior performance-bandwidth trade-offs under different communication constraints, demonstrating its ability to adjust to various communication limitations and effectively optimize communication efficiency in different scenarios; and iii) PragComm achieves more than $32.7$K lower communication volume while still outperforms previous state-of-the-art collaborative methods.

The contributions of this work are as follows:

$\bullet$ We propose a novel pragmatic communication strategy that only transmits task-critical data, reducing communication cost while retaining task utility.


$\bullet$ We propose a pragmatic collaborative perception system that leverages pragmatic messages to transmit task-critical data and subsequently integrates them to enhance collaborative perception capabilities.



$\bullet$ We conduct extensive experiments on multiple perception tasks to validate the superiority of PragComm on the performance-communication trade-off.

\vspace{-2mm}
\section{Related Work}
\label{sec:related work}

\textbf{Collaborative perception.}
Collaborative perception is an application of multi-agent systems. It addresses the inherent limitations in single-agent perception, like occlusion and long-range challenges, by allowing agents to exchange supplemental information. This emerging field has seen a wave of premium datasets to boost algorithmic development. Some of these include simulated datasets like V2X-SIM~\cite{LiV2XSim:RAL22}, OPV2V~\cite{XuOPV2V:ICRA22}, and CoPerception-UAVs~\cite{HuWhere2comm:NeurIPS22}, as well as real-world datasets like DAIR-V2X~\cite{YuDAIRV2X:CVPR22} and V2V4Real~\cite{XuV2V4Real:CVPR23}.
Collaborative perception systems have made remarkable progress in improving perception performance and robustness. 
In terms of perception performance, studies like ~\cite{XuOPV2V:ICRA22,XuCoBEVT:CoRL22,XuV2XViT:ECCV22} introduce transformer structures for more streamlined data aggregation. DiscoNet~\cite{LiLearning:NeurIPS21} expands fusion scalars to matrices with spatial dimensions, allowing for region-specific information amalgamation.
In terms of robustness, SyncNet~\cite{LeiLatency:ECCV22} offers a method to counteract communication delay, and CoAlign~\cite{LuRobust:ICRA23} uses an agent-object pose graph to address pose inaccuracies.
Here we mainly focus on communication-constrained collaborative perception. Instead of solely promoting the perception performance without evaluating the expense of communication bandwidth, we aim to optimize the performance-communication trade-off.


\textbf{Communication efficiency in collaborative perception.}
Communication efficiency is the bottleneck for the scale-up of collaborative perception, as real-world communication resource is always constrained and can hardly support huge communication consumption in real-time. 
Previous works can be categorized into three types: i) early collaboration~\cite{ChenCooper:ICDCS19}, which transmits raw observation data. While this preserves the most complete information, it consumes considerable bandwidth; ii) late collaboration, which transmits detected boxes. This is bandwidth-efficient but offers sub-optimal perception performance, as it fails to rectify any missed detections in a single-agent view; and iii) intermediate collaboration~\cite{LiuWhen2com:CVPR20,LiuWho2com:ICRA20,WangV2vnet:ECCV20,HuWhere2comm:NeurIPS22,YangWhat2comm:ICCV2023,WangUMCAU:CVPR2023,How2comm:NeurIPS2023}, which transmits representative information with compact features, balancing communication bandwidth and perception performance.

Intermediate collaboration enhances communication efficiency in various ways. V2VNet~\cite{WangV2vnet:ECCV20} and DiscoNet~\cite{LiLearning:NeurIPS21} use a fully-connected communication graph and utilize source coding~\cite{balle2018variational} and autoencoders~\cite{masci2011stacked} to represent feature maps more efficiently. When2com~\cite{LiuWhen2com:CVPR20} refines the dense communication graph to a sparser one using a handshake mechanism.
However, past efforts assume collaborating agents consistently share all perception data across time and space. This can be wasteful, as many regions and moments might offer little value to perception tasks.
Addressing this, we adopt a perception-centric pragmatic communication strategy, substantially eliminating task-irrelevant data.

\textbf{Pragmatic compression.}
Pragmatic compression focuses on capturing key information and creating compact representation for specific downstream tasks. Rather than prioritizing perfect data reconstruction, this approach emphasizes extracting information that best serves the intended tasks.
Singh~\cite{singh2020end} considers image classification and designs the loss function to balance task accuracy and bit rate. Dubois~\cite{dubois2021lossy} designs representations tailored for multiple downstream tasks, considering the assumption that they favor the transformation-invariant information of inputs. Reddy~\cite{reddy2021pragmatic} employs adversarial training to ensure the original and compressed inputs lead to consistent outcomes. However, all previous works focus on classification tasks and solely shrink the channel dimension. Here, we address the intricate tasks of object detection and tracking. Considering the perceptual objective, we achieve a comprehensive compression across spatial, temporal, and channel dimensions, significantly boosting communication efficiency.

\textbf{Semantic communication.} 
Semantic communication~\cite{QinSemantic:ArXiv21,KalfaTowards:DSP21,XieTaskOriented:WCL21,XieALite:JSAC20,WengSemantic:JSAC21,XieDeep:TSP20,xie2021deep, qin2021semantic, luo2022semantic} aims at effective semantic information exchange rather than accurate symbol transmission regardless of its meaning.
Efforts to realize this vision have spanned several aspects.
In the realm of semantic entropy, previous studies~\cite{FarsadDeep:ICASSP18,XieDeep:TSP20} have measured it using the degree of confirmation, an approach inspired by Shannon's foundational concept of information entropy.
Regarding the design of semantic communication systems, previous work~\cite{LanWhat:Arxiv21} has incorporated the semantic layer into conventional communication systems, leading to the creation of a semantic open system interconnection model.
In the field of source-coding technologies, previous works~\cite{CarnapAnOutline:RLE52,BaoTowards:TNSW11} have developed a transformer-based joint semantic-channel coding mechanism, culminating in an end-to-end semantic communication system optimized for text and speech data.
For the formulation of evaluation metrics suited to various sources such as text, images, and speech, previous works~\cite{LahatMultimodal:PIEEE15,KurkaDeepJSCC:JSAIT19,KurkaBandwidthAgile:ITWC20} have designed adaptive-bandwidth wireless transmission for images and multi-media data.  
Despite these advancements, existing works are all about the isolated exploration of semantic communication within the confines of a singular sender-receiver relationship, with a predominant focus on semantic representation. In contrast, our work considers semantic communication in a multi-agent collaboration setting. We put forth a pragmatic communication strategy that comprehensively addresses semantic information representation, communication graph construction, and message fusion across multiple agents.

\vspace{-2mm}
\section{Collaborative Perception Optimization}
\label{sec:formulation}

In this section, we present collaborative perception as a constrained optimization problem. We first formulate the communication-constrained collaborative perception objective, this is, maximizing perception performance while respecting the communication budget constraint, in Section~\ref{sec:objective}. To solve this, we propose a novel pragmatic communication strategy in Section~\ref{sec:strategy}. Based on this strategy, we reformulate the optimization and propose a solution in Section~\ref{sec:optimization}.


\vspace{-2mm}
\subsection{Perception-communication trade-off}\label{sec:objective}
The overall objective is to maximize the perception performance over a time duration of all agents by exchanging complementary perceptual information among them while staying within the total communication budget; that is, optimizing the perception-communication trade-off. Here, we consider the perception task of object detection and tracking over a time duration of $T$. Consider $N$ agents in the scene, let $\mathcal{X}_i^{(t)}$ and $\widehat{\mathcal{T}_{i}}^{(t)}$ be the observation and the target of the $i$th agent at the time stamp $t$, respectively, and $B$ be the total communication budget. 
Then, the objective is formulated as
\begin{subequations}
\setlength\abovedisplayskip{1pt}
\setlength\belowdisplayskip{1pt}
\begin{align}
\underset{\theta,\mathcal{P}}{\arg \max}~\sum_{i=1}^{N} g (\Phi_{\theta} &\left(\{\mathcal{X}_i^{(t)}\}_{t=1}^T,\{\mathcal{P}_{j\rightarrow i}^{(t)}\}_{j=1,t=1}^{N,T} \right), 
\{\widehat{\mathcal{T}_{i}}^{(t)}\}_{t=1}^{T}),\label{eq:objective}\\[-10pt]
\text{s.t.}~ &\sum_{t=1}^{T} \sum_{i,j=1,i\neq j}^{N}b({\mathcal{P}_{j\rightarrow i}^{(t)}}) \leq B,\label{eq:constraint}
\end{align}    
\end{subequations}
where $g(\cdot,\cdot)$ is the non-differentiable perception evaluation metric which measures the difference between the perception prediction and the ground truth, for instance, average precision (AP) in terms of detection task, and multi-object tracking accuracy (MOTA) in terms of tracking task. And $\mathcal{P}_{j\rightarrow i}^{(t)}$ is the collaborative message transmitted from the $j$th agent to $i$th agent at the $t$th timestamp, and the message when $j=i$ equals to the encoded observation and does not incur any communication cost, $b(\cdot)$ is the non-differentiable bandwidth cost quantification metric which measures the communication bits of the pragmatic messages, ${\Phi}_{\theta}$ is the collaborative perception network with trainable parameter $\theta$, which transforms the individual observations and the received pragmatic messages into perception results.

It is extremely challenging to optimize the objective. The reasons are i) due to the hard budget constraints in Equation~\eqref{eq:constraint} and non-differentiable task-utility measurement in Equation~\eqref{eq:objective}, it is hard to directly optimize this objective; and ii) due to the diverse data inputs and numerous network parameters needed for complex task functionality, it is hard to enumerate the large search space to find the solution.
To solve this objective, we propose a novel pragmatic communication strategy in Section~\ref{sec:strategy}.



\vspace{-3mm}
\subsection{Pragmatic communication strategy}
\label{sec:strategy}
The key to solving the objective lies in optimizing messages to fill each agent's specific perception task demand within the communication budget. 
To achieve this, the pragmatic communication strategy crafts pragmatic messages from the complete input data from three aspects: message selection (\textit{what to communicate}), message representation (\textit{how to communicate}), and collaborator selection (\textit{who to communicate}).
Specifically, the pragmatic messages are given by 
\begin{equation}
    \mathcal{P}=\Psi_{\rm who}\left(\Psi_{\rm represent}\left(\Psi_{\rm select}\left(\mathcal{X}\right)\right)\right).
\end{equation}
First, pragmatic message selection $\Psi_{\rm select}(\cdot)$ identifies informative regions essential for the task objective. For instance, in question-answering tasks, key entities are crucial, while in segmentation and classification tasks, boundary regions and foreground areas are more important, respectively. Second, pragmatic message representation $\Psi_{\rm represent}(\cdot)$ employs efficient representation to convey the task-essential information within the selected regions. For instance, in occupancy prediction tasks, detailed information such as color, and structures are less important. Third, pragmatic collaborator selection $\Psi_{\rm who}(\cdot)$ selects beneficial collaborators to exchange these information representations. For instance, in detection tasks, the communication link between agents who have the same observations can be cut off, as they can not provide complementary information to each other.

Here, we mainly focus on the collaborative perception task of detection and tracking where each agent has its specific objective to identify the foreground object sequences over time within its designated perception region. 
First, in pragmatic message selection, we treat the targeted foreground object sequences as sparse spatial-temporal signals. This means we only select sparse spatial regions containing foreground objects and sample the critical temporal moments that capture their dynamics for our pragmatic messages.
Second, regarding pragmatic message representation, we draw inspiration from efficient dictionary-based human communication. We standardize agent communication by maintaining a common dictionary of perceptual features. This enables us to approximate the original high-dimensional feature vectors with compact indices, leading to more efficient message representation.
Third, in pragmatic collaborator selection, considering each agent's unique objective, we involve cross-temporal collaboration to iteratively refine each collaborator’s needs and accordingly provide the demanded data to the beneficial collaborator.
Furthermore, owing to the demonstrated success of feature-based intermediate collaboration in prior research~\cite{LiLearning:NeurIPS21,XuV2XViT:ECCV22}, our pragmatic communication strategy collaborates with intermediate features.



Based on the above intuition, the pragmatic communication strategy implements i) pragmatic message selection by employing a binary spatial selection matrix $\mathbf{S}_{i}^{(t)}$ and a binary temporal selection matrix $\mathbf{T}_{i}^{(t)}$ to pick out the observable foreground regions for the supporter $i$, and a binary spatial selection matrix $\overline{\mathbf{S}}_{j}^{(t-1)}$ to reflect the collaborator $j$'s pragmatic demands. Then the element-wise multiplication of these matrix $\mathbf{S}_i^{(t)}\odot\mathbf{T}_{i}^{(t)}\odot\overline{\mathbf{S}}_j^{(t-1)}$ jointly selects the necessary spatial regions where the supporter can provide and the collaborator demands at each timestamp; ii) pragmatic message representation by constructing a common task-adaptive codebook $\mathbf{D}$ and then employ a pragmatic approximation function $\Phi_{\mathbf{D}}(\cdot)$ to replace the original perceptual feature with the nearest code indices; and iii) pragmatic collaborator selection by employing a binary adjacency matrix $\mathbf{A}^{(t)}\in \{0, 1\}^{N\times N}$, where each scalar $\mathbf{A}^{(t)}_{i,j}$ at position $(i,j)$ indicates message passing from the $i$th agent to the $j$th agent and is set $1$ only when the collaborator's demanded information is present in the message. 

The pragmatic communication strategy implements the message $\mathcal{P}_{j\rightarrow i}^{(t)}$ transmitted from the $j$th agent to $i$th agent at the $t$th timestamp with 
\begin{equation}
    \mathcal{P}_{i\rightarrow j}^{(t)} = \mathbf{A}^{(t)}_{i,j}\Phi_{\mathbf{D}}(\Phi_{\rm enc}(\mathcal{X}_{i}^{(t)})\odot\mathbf{S}_i^{(t)}\odot\mathbf{T}_{i}^{(t)}\odot\overline{\mathbf{S}}_j^{(t-1)}),\label{eq:message}
\end{equation}
where $\Phi_{\rm enc}(\cdot)$ is the observation encoder extracts the intermediate feature map from the input $\mathcal{X}_{i}^{(t)}$.
Note that collaboration is not isolated at each timestamp but is continually influenced by evolving cross-temporal demands.
Here, $\overline{\mathbf{S}}_{j}^{(t-1)}$ reflects collaborator $j$'s demand, derived as the complement of $\mathbf{S}_{j}^{(t-1)}$, this is, $\overline{\mathbf{S}}_{j}^{(t-1)}=1-\mathbf{S}_{j}^{(t-1)}$, signifying non-selected regions in previous communication round. The key idea is that these non-selected regions correspond to areas in the collaborator's view without objects, potentially containing missed objects due to limited visibility such as occlusion. Thereby collaborator demands additional information in these seemingly empty areas, which may recover missed objects, improving perception accuracy.


To sum up, Equation~\eqref{eq:message} achieves a valid pragmatic communication strategy by using $\mathbf{S}_i^{(t)}\odot\mathbf{T}_{i}^{(t)}\odot\overline{\mathbf{S}}_j^{(t-1)}$ to select the spatial-temporal regions containing task-critical foreground object dynamics where supporter can provide while the collaborator requires, $\Phi_{\mathbf{D}}(\cdot)$ to represent the selected information with concise code index, and $\mathbf{A}^{(t)}_{i,j}$ to transmit these pragmatic messages to the beneficial collaborator.

\vspace{-3mm}
\subsection{Reformulated optimization}
\label{sec:optimization}

Here based on the pragmatic messages in Equation~\eqref{eq:message}, we reformulate the objective Equation~\eqref{eq:objective} in Section~\ref{sec:objective} as
\begin{subequations}
\setlength\abovedisplayskip{1pt}
\setlength\belowdisplayskip{1pt}
\begin{align}
\underset{\theta,\mathbf{A},\mathbf{S},\mathbf{T},\mathbf{D}}{\arg \max}~\sum_{i=1}^{N} g &(\Phi_{\theta} \left(\{\mathcal{X}_i^ {(t)}\}_{t=1}^T,\{\mathcal{P}_{j\rightarrow i}^{(t)}\}_{j=1,t=1}^{N,T} \right), \{\widehat{\mathcal{T}_{i}}^{(t)}\}_{t=1}^{T} ),\label{subeq:objective}\\[-10pt]
   \text{s.t.}~~\mathcal{P}_{i\rightarrow j}^{(t)}= \mathbf{A}^{(t)}_{i,j}&\Phi_{\mathbf{D}}(\Phi_{\rm enc}(\mathcal{X}_{i}^{(t)})\odot\mathbf{S}_i^{(t)}\odot\mathbf{T}_{i}^{(t)}\odot\overline{\mathbf{S}}_j^{(t-1)}),\label{subeq:message}\\[-10pt]
   ~~\sum_{t=1}^{T} \sum_{i=1,j\neq i}^{N} &\mathbf{A}^{(t)}_{i,j}|\mathbf{S}_i^{(t)}\odot\mathbf{T}_j^{(t)}\odot\overline{\mathbf{S}}_j^{(t-1)}| n_D \leq B.\label{subeq:constraint}
\end{align}
\end{subequations}
The pragmatic communication strategy establishes an optimizable connection between the non-differentiable task utility in~\eqref{subeq:objective} and the hard communication constraint in~\eqref{subeq:constraint} by reformulating message $\mathcal{P}$ with three optimizable key factors as outlined in~\eqref{subeq:message}: 
i) optimizing spatial and temporal selection matrix $\mathbf{S}$ and $\mathbf{T}$ for pragmatic message selection, reaching optimal when all areas containing object dynamics are selected, ii) optimizing codebook $\mathbf{D}$ for pragmatic message representation, reaching optimal when all essential perceptual bases are preserved; and iii) optimizing adjacent matrix $\mathbf{A}$ for pragmatic collaborator selection, reaching optimal when all collaborators who benefit from the supporter's observable targets are selected.
These optimizable parameters enable both the flexible selection and full preservation of task-critical information, optimizing the efficiency while retaining the optimal solution with the original problem in~\eqref{eq:objective}.

Furthermore, to solve this reformulated objective in Equation~\eqref{subeq:objective} with hard constraint in Equation~\eqref{subeq:constraint}, we decompose it into two sub-optimization problems and alternatively optimize the efficiency of the collaborative message and the network parameters once at a time: i) given the feasible perception network parameter, obtain efficient and informative pragmatic messages $\mathcal{P}$; ii) given the feasible pragmatic messages, optimize the collaborative perception network parameter $\theta$. The constraint is satisfied in i) and the perception goal is achieved in ii). 
Note that this problem-solving heuristic follows the alternating direction method of multipliers (ADMM)~\cite{ADMM} tactic. The two sub-optimizations can be iteratively optimized to increasingly approach the optimal solution. In this study, we employ just one iteration due to the efficiency and real-time demands. The results demonstrate that a single iteration is indeed viable. 


\vspace{-2mm}
\subsubsection{Message determination}
\label{sec:determination}

Here, we optimize the first sub-optimization problem, this is, determining feasible pragmatic messages $\mathcal{P}$ that contain the most critical perception information while satisfying the communication constraint. To achieve this, we initialize the perception network with a feasible parameter $\theta$, sourced from the optimized solution of single-agent perception. Then this sub-problem of Equation~\eqref{subeq:objective} is given by
\begin{subequations}
\setlength\abovedisplayskip{1pt}
\setlength\belowdisplayskip{1pt}
\begin{align}
\underset{\mathbf{A},\mathbf{S},\mathbf{T},\mathbf{D}}{\arg \max}~\sum_{i=1}^{N} g &(\Phi_{\theta} \left(\{\mathcal{X}_i^ {(t)}\}_{t=1}^T,\{\mathcal{P}_{j\rightarrow i}^{(t)}\}_{j=1,t=1}^{N,T} \right), \{\widehat{\mathcal{T}_{i}}^{(t)}\}_{t=1}^{T} ),\label{subeq:ori_objective}\\[-10pt]
   \text{s.t.} ~\mathcal{P}_{i\rightarrow j}^{(t)}= \mathbf{A}^{(t)}_{i,j}&\Phi_{\mathbf{D}}(\Phi_{\rm enc}(\mathcal{X}_{i}^{(t)})\odot\mathbf{S}_i^{(t)}\odot\mathbf{T}_{i}^{(t)}\odot\overline{\mathbf{S}}_j^{(t-1)}),\label{subeq:opt_message}\\[-10pt]
   \sum_{t=1}^{T} \sum_{i=1,j\neq i}^{N} &\mathbf{A}^{(t)}_{i,j}|\mathbf{S}_i^{(t)}\odot\mathbf{T}_i^{(t)}\odot\overline{\mathbf{S}}_j^{(t-1)}| n_D \leq B.\label{subeq:opt_constraint}
\end{align}
\end{subequations}
The feasible perception network serves as an evaluation function, which enables the assessment of the task utility given the obtained pragmatic messages. This guides us to prioritize the most perceptually critical information in each criterion: information selection, information representation, and collaborator selection by optimizing $\mathbf{S}$, $\mathbf{T}$, $\mathbf{D}$, and $\mathbf{A}$, respectively, while adhering to bandwidth constraints $B$. The implementation details can be referred to in Section~\ref{sec4:determination}.

\vspace{-2mm}
\subsubsection{Message utilization}
\label{sec:utilization}

Here, we optimize the second sub-optimization problem, this is, obtaining the optimal network parameter $\theta$ that maximize the perception performance given the optimized pragmatic messages $\mathcal{P}$ in the previous sub-optimization~\eqref{subeq:ori_objective}. Specifically, this sub-problem of~\eqref{subeq:objective} is given by
\begin{subequations}
\setlength\abovedisplayskip{-1pt}
\setlength\belowdisplayskip{-1pt}
\begin{align}
    \underset{\theta}{\min}~\sum_{t=1}^{T}\sum_{i=1}^{N} 
    L_{\text{det}} \left(\Phi_{\theta} \left(\{\mathcal{X}_i^{(k)}\}_{k=1}^{t},\{\mathcal{P}_{j\rightarrow i}^{(k)}\}_{j=1,k=1}^{N,t}
    \right), \widehat{\mathcal{T}_{i}}^{(t)} \right).~\label{subeq:perfromance_subobjective}
\end{align}    
\end{subequations}
We separate the time series objective and use a greedy approach to minimize the perception evaluation metric for each timestamp. The collaborative perception network optimizes the use of all the available pragmatic messages up to each timestamp $t$, including both fresh and historical ones to offer spatial-temporal complementarity.
The perception evaluation metric $g(\cdot)$ is achieved with the detection loss $L_{\rm det}$, which calculates the difference between predictions with the ground truth. Note that this sub-problem~\eqref{subeq:perfromance_subobjective} does not involve any constraints, making it straightforward to be solved by standard~\textbf{supervised learning}. The implementation details can be referred to in Section~\ref{sec4:utilization}.

\textbf{Advantages of pragmatic communication strategy.} 
Specifically, pragmatic communication strategy significantly reduces communication costs in three aspects:
i) pragmatic message selection reduces the communication cost from $T\cdot H\cdot W$ to $|\mathbf{S}\odot\mathbf{T}\odot\overline{\mathbf{S}}|$ by prioritizing foreground object dynamics;
ii) pragmatic message representation cuts the communication cost from $C$ to $n_D$ by replacing high-dimensional feature vectors with pragmatic approximation achieved with code indices;
and iii) pragmatic collaborator selection trims the communication cost from $N\cdot N$ to $|\mathbf{A}|$ by focusing on the beneficial collaborators.
Moreover, the experimental results in Fig.~\ref{Fig:ABL_STC} demonstrate that the communication efficiency is substantially enhanced with these pragmatic designs, validating that the proposed criteria effectively measure the task critical level and eliminate orthogonal redundancy.

In the next section, we propose a novel pragmatic collaborative perception system, which implements the pragmatic communication strategy with learnable neural networks and leverages these pragmatic messages to enhance detection and tracking capabilities of multiple collaborative agents.

\section{Pragmatic Collaborative Perception System}
\label{sec:system}
\vspace{-1mm}

In this section, we propose a novel neural network-based implementation for the pragmatic communication strategy in Section~\ref{sec:formulation}.
Our proposed pragmatic collaborative perception system, PragComm, comprises two key components: 
i) a unified single-agent perception system that achieves detection and tracking in Section~\ref{sec4:single_perception};
and ii) multi-agent pragmatic collaboration that achieves collaborative message determination in Section~\ref{sec4:determination}, and utilization in Section~\ref{sec4:utilization}.

The message determination implements the first sub-optimization~\eqref{subeq:ori_objective} in Section~\ref{sec:determination}, determining the collaborative message $\mathcal{P}$. The message utilization realizes the second sub-optimization~\eqref{subeq:perfromance_subobjective} in Section~\ref{sec:utilization}, utilizing the collaborative message to enhance perception performance.


\vspace{-1mm}
\subsection{Single-agent detection and tracking}
\label{sec4:single_perception}

The unified perception system achieves both 3D detection and tracking. and comprises three essential components: observation encoder, detection decoder, and tracker. The observation encoder extracts 3D features from raw sensor inputs. The detection decoder takes the intermediate features and decodes them into objects, represented by bounding boxes. Finally, the tracker associates objects across different timestamps, forming object sequences.

\vspace{-1mm}
\subsubsection{Observation encoder}
\label{sec4:encoder}
The observation encoder extracts feature maps from the sensor data. It accepts single/multi-modality inputs, such as RGB images and 3D point clouds. For the $i$th agent at timestamp $t$, given its input $\mathcal{X}_i^{(t)}$, the extracted feature map is 
$ 
\mathcal{F}_i^{(t)} = \Phi_{\rm enc}(\mathcal{X}_i^{(t)}) \in \mathbb{R}^{H \times W \times C},$
where $\Phi_{\rm enc}(\cdot)$ is the encoder, and $H,W,C$ are its height, weight and channel. 
Note that this work adopts the feature representations in bird's eye view (BEV), where all agents project their individual perceptual information to the same global coordinate system, avoiding complex coordinate transformations and supporting better shared cross-agent collaboration. 
For the image input, $\Phi_{\rm enc}(\cdot)$ is followed by a warping function that transforms the extracted feature from front-view to BEV. For 3D point cloud input, we discretize 3D points as a BEV map and $\Phi_{\rm enc}(\cdot)$ extracts features in BEV.

\vspace{-1mm}
\subsubsection{Detection decoder}
\label{sec4:decoder}
The detection decoder takes the BEV feature as input and outputs the objects, including class and regression. 
Given the feature map at the $t$th timestamp $\mathcal{F}_i^{(t)}$, the detection decoder $\Phi_{\rm dec}(\cdot)$ generate the dense heatmap of $i$th agent by
$
     \mathcal{O}_i^{(t)}  =  \Phi_{\rm dec}(\mathcal{F}_i^{(t)}) \in \mathbb{R}^{H \times W \times 7},
$
where each location of $\mathcal{O}_i^{(t)}$ represents a rotated box with class $(c,x,y,h,w, \cos\alpha, \sin\alpha)$, denoting class confidence, position, size and angle. Non-maximum suppression (NMS) is applied to the dense predictions and generates the sparse output of the 3D detection system by $\mathcal{O'}_i^{(t)}=\Phi_{\rm NMS}\left(\mathcal{O}_i^{(t)}\right)\in\mathbb{R}^{M_i^{(t)}\times 7}$, where $M_i^{(t)}$ is the final number of objects in the sparse output.
\vspace{-1mm}
\subsubsection{Tracker}
\label{sec4:tracker}
The tracker associates 3D detections with coherent trajectories. At each timestamp, to associate the current detections with the historical trajectories, it requires three essential steps: i) \textbf{prediction}, where a 3D Kalman filter predicts the state of existing trajectories from previous timestamps to the current timestamp; ii) \textbf{data association}, which matches the detections in the current timestamp with the nearest predicted trajectories; and iii) \textbf{update}, where the 3D Kalman filter updates the state of matched trajectories and creates new trajectories for newly detected objects or removes trajectories for disappeared objects.
To perform the prediction from historical states to the current timestamp, we use a constant velocity model to approximate the objects' displacement. Each trajectory state is represented as a $9$-dimensional vector $(c,x,y,h,w,\cos\alpha,\sin\alpha,v_x,v_y)$, where the additional variables $v_x$ and $v_y$ denote the object's velocity in the bird's eye view (BEV) space.
Specifically, for agent $i$ at timestamp $t$, given the $\overline{M}_i^{(t-1)}$ historical trajectories with state $\mathcal{T}_{i}^{(t-1)}\in\mathbb{R}^{\overline{M}_i^{(t-1)}\times 9}$ and the $M_i^{(t)}$ current sparse detections $\mathcal{O'}_i^{(t)}\in\mathbb{R}^{M_i^{(t)}\times 7}$, the tracking process is given by:
\begin{subequations}
\begin{align}
\overline{\mathcal{T}}_{i}^{(t)}&=\Phi_{\rm Kalman,P}\left(\mathcal{T}_{i}^{(t-1)}\right)\in\mathbb{R}^{\overline{M}_i^{(t-1)}\times 9},\label{eq:kalman_pred} \\ 
\mathbf{G}_i^{(t)}&=\Phi_{\rm Kalman,A}\left(\overline{\mathcal{T}}_{i}^{(t)},\mathcal{O'}_i^{(t)}\right)\in\{0,1\}^{\overline{M}_i^{(t-1)}\times M_i^{(t)}},\label{eq:kalman_associate} \\ 
\mathcal{T}_{i}^{(t)}&=\Phi_{\rm Kalman,U}\left(\mathcal{T}_{i}^{(t-1)},\mathbf{G}_i^{(t)},\mathcal{O'}_i^{(t)}\right)\in\mathbb{R}^{\overline{M}_i^{(t)}\times 9},\label{eq:kalman_update}
\end{align}    
\end{subequations}
where $\Phi_{\rm Kalman,P}(\cdot)$, $\Phi_{\rm Kalman,A}(\cdot)$, and $\Phi_{\rm Kalman,U}(\cdot)$ represent the trajectory prediction, data association, and trajectory update functions of the 3D Kalman filter $\Phi_{\rm Kalman}(\cdot)$. Note that, the implementation of the tracker follows~\cite{WengAB3DMOT:IROS2020}. 

Equation~\eqref{eq:kalman_pred} predicts the trajectory $\overline{\mathcal{T}}_{i}^{(t)}$ at the current timestamp based on the constant velocity model, where $x_{\rm est}=x+v_x$ and $y_{\rm est}=y+v_y$.
Equation~\eqref{eq:kalman_associate} calculates the binary association matrix $\mathbf{G}_i^{(t)}$ of size $\overline{M}_i^{(t-1)}\times M_i^{(t)}$. Each element reflects whether the predicted track and the object detection are matched or not, where $1$ denotes a match and $0$ denotes no match. For the matching process, we adopt the 3D Intersection over Union (IoU) as the matching score. The intuition is to associate each detection with the most overlapped historical trajectories.
Note that the matching process involves two steps: i) generating an affinity matrix by computing IoUs between $\overline{\mathcal{T}}_{i}^{(t)}$ and the object detections $\mathcal{O'}_i^{(t)}$, and ii) get the binary matching matrix by solving the bipartite graph matching problem in polynomial time using the Hungarian algorithm.
Equation~\eqref{eq:kalman_update} updates the tracks $\mathcal{T}_{i}^{(t)}$ based on the association matrix $\mathbf{G}_i^{(t)}$ and the object detections $\mathcal{O'}_i^{(t)}$ using the 3D Kalman filter.
As tracked objects may leave the scene and new objects may enter the scene, it is necessary to manage the birth and death of trajectories. The updating process has three cases: i) for the matched historical trajectories, the trajectory state is updated using the Bayes update rule; ii) for the unmatched historical trajectories, the death rule is applied to determine if the trajectory should be terminated; and iii) for the unmatched detections, a new trajectory is created and initialized with the detection, assuming zero velocity for $v_x$ and $v_y$.

This subsection presents a single-agent detection and tracking system. However, single-agent perception systems face inevitable challenges, such as long-range perception and occlusions. To overcome these limitations, we achieve a pragmatic collaboration solution in the following sections, which allows agents to exchange informative cues and compensate for each other's shortcomings through communication. 


\begin{figure*}[!t]
    \centering
    \includegraphics[width=1.0\linewidth]{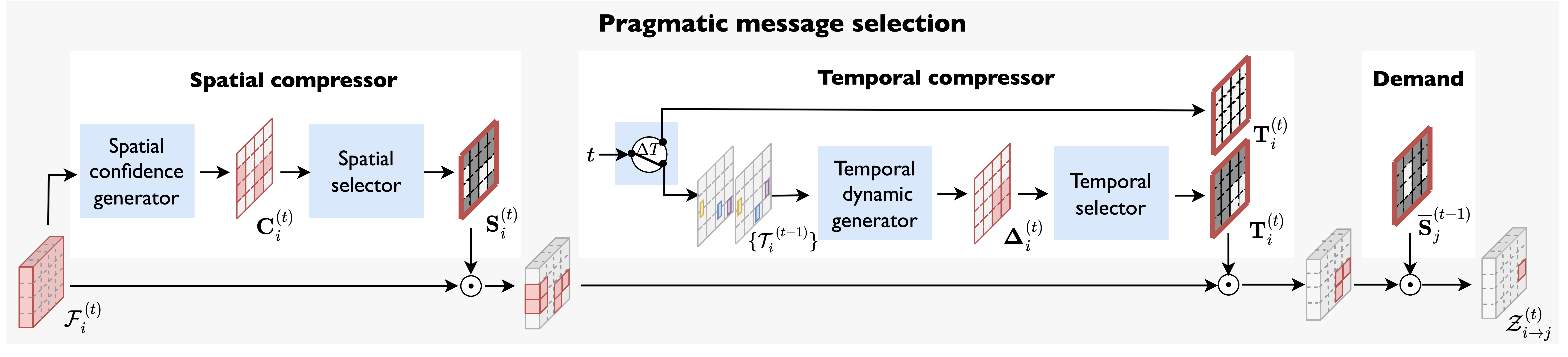}
    \vspace{-7mm}
    \caption{Overview of the spatial and temporal compressor. The spatial compressor picks out the perceptual critical foreground regions. The temporal compressor has two options: when reaching the updating frequency it selects all regions (the upper branch), or it picks out the dynamic regions (the bottom branch).}
    \vspace{-6mm}
    \label{fig:spatial}
\end{figure*}

\subsection{Pragmatic collaboration: Message determination}
\label{sec4:determination}
Referring to the analysis in Section~\ref{sec:formulation}, this subsection implements the first sub-optimization~\eqref{subeq:ori_objective}, determining the collaborative message $\mathcal{P}$ within the communication budget. To achieve this, we implement each element of $\mathcal{P}$ in Equation~\eqref{subeq:opt_message} with neural networks.

For pragmatic message selection, the spatial selection matrix $\mathbf{S}$ is achieved with a spatial compressor in Section~\ref{sec4:where2comm}, and the temporal selection matrix $\mathbf{T}$ is achieved with a temporal compressor in Section~\ref{sec4:when2comm}.
These matrices specify the available information from each supporter and the demanded information from each collaborator, and they together determine the pragmatic information.
For pragmatic message representation, the codebook $\mathbf{D}$ and function $\Phi_{\mathbf{D}}(\cdot)$ is achieved with a channel compressor in Section~\ref{sec4:what2comm}.
For pragmatic collaborator selection, the collaborator selection matrix $\mathbf{A}$ is achieved in message exchange in Section~\ref{sec4:exchange}.

\vspace{-1mm}
\subsubsection{Pragmatic message selection: Spatial compressor}
\label{sec4:where2comm}

The spatial compressor implements pragmatic message selection in the spatial aspect by selectively communicating at foreground areas containing objects rather than the full spatial region.
The intuition is that during collaboration the perceptual information in these foreground areas can help recover the objects occluded in the other's views while the background regions can be omitted to promote communication efficiency. 
It obtains a locally optimal binary spatial selection matrix $\mathbf{S}$ in Equation~\eqref{subeq:opt_message}.



To achieve this, the spatial compressor includes two phases: i) generate a spatial confidence map $\mathbf{C}_{i}^{(t)}$ that reflects the perceptual critical level, this is, the possibility that a spatial region contains objects, and ii) obtain a locally optimal binary spatial selection matrix $\mathbf{S}$ in~\eqref{subeq:opt_message} by leveraging a selection function $\Phi_{\rm select}(\cdot,\cdot)$ to prioritize the perceptually critical spatial regions to be included in the message. 

\textbf{Spatial confidence map generation.} Here, we implement the spatial confidence map with the detection confidence map, which accurately reflects the spatial heterogeneity of perceptual information, where the area with a high perceptually critical level is the area that contains an object with a high confidence score. Given the perceptual feature $\mathcal{F}_i^{(t)}$ at $t$th timestamp, its spatial confidence map $\mathbf{C}_i^{(t)}$ is 
\begin{equation}
\label{eq:generator}
    \mathbf{C}_i^{(t)}  =  \Phi_{\rm dec}(\mathcal{F}_i^{(t)})  \in [0,1]^{H \times W}.
\end{equation}
We iteratively update the spatial confidence map at each timestamp. Once $\mathcal{F}_i^{(t)}$ is obtained, Equation~\eqref{eq:generator} is triggered to reflect the perceptually critical level at each spatial location. The proposed spatial confidence map answers a crucial question that was ignored by previous works: for each agent, information at which spatial area is worth sharing with others. By answering this, it provides a solid base for efficient communication.

\textbf{Spatial selection.} Here, we obtain a binary selection matrix to determine what information can be omitted in the full feature map and pack the spatially sparse, yet perceptually critical feature map into the to-be-sent messages. We implement $\Phi_{\rm select}(\cdot)$ by selecting the most critical areas conditioned on the spatial confidence maps and the bandwidth limit. Specifically, we achieve this selection function as the solution of a proxy-constrained problem as follows,
\begin{equation}
    \underset{\mathbf{S}}{\max}~ \mathbf{S}_{i}^{(t)} \odot \mathbf{C}_i^{(t)},~~{\rm s.t.~} |\mathbf{S}_{i}^{(t)}| \leq b_i, ~~ \mathbf{S}_{i}^{(t)}\in\{0,1\}^{H\times W},
\end{equation}
where $\odot$ is the element-wise multiplication, the optimal selection matrix $\mathbf{S}_{i}^{(t)}$ selects the locations where the largest elements in the given input matrix conditioned on the bandwidth limit $b_i$. Note that even though this optimization problem has hard constraints and non-differentialability of binary variables, it has an analytical solution that satisfies the constraint. This solution is obtained by selecting those spatial regions whose corresponding elements are in $\mathbf{S}$ rank top-$b_i$. The detailed steps of \textbf{selection function} are: i) arrange the elements in the input matrix in descending order; ii) given the communication budget constrain, decide the total number $b_i$ of communication regions; iii) set the spatial regions of $\mathbf{S}$, where elements rank in top-$b_i$ as the $1$ and $0$ verses.

\vspace{-1mm}
\subsubsection{Pragmatic message selection: Temporal compressor}
\label{sec4:when2comm}

The temporal compressor implements pragmatic message selection in the temporal aspect by selectively communicating at the essential timestamps capturing object dynamics rather than the full-time duration.
The intuition is to treat object trajectories as temporal signals and leverage the Nyquist Sampling Theorem, which states that a sampling frequency twice the highest frequency in the temporal sequence is adequate for accurate signal reconstruction. By sampling key timestamps, we ensure to capture essential dynamics, reducing the amount of communication needed while retaining crucial information. 
It obtains the locally optimal temporal sampling matrix $\mathbf{T}$ in Equation~\eqref{subeq:opt_message}.

To achieve this, the temporal compressor includes two phases: i) generate a dynamic matrix $\mathbf{\Delta}_{i}^{(t)}$ that reflects the object dynamic level, with higher scores indicating regions where objects exhibit greater changes in their motion patterns between the most recent timestamps, and ii) obtain a locally optimal binary temporal selection matrix $\mathbf{T}$ in~\eqref{subeq:opt_message} by leveraging a sampling function $\Phi_{\rm sample}(\cdot,\cdot)$ to prioritize the regions that capturing essential object dynamics to be included in the message.

\textbf{Temporal dynamic estimation.} Here, we achieve the dynamic estimator with the 3D Kalman filter $\Phi_{\rm Kalman}(\cdot)$ in the tracker (Section~\ref{sec4:tracker}). The object dynamic matrix is obtained by calculating the difference of object states between the most recent two timestamps. The intuition is that the object's dynamic level is reflected in the changes in its state from various perspectives, particularly in terms of velocity. Specifically, the dynamic matrix $\mathbf{\Delta}_i^{(t)}$ is given by 
\begin{equation}
\label{subeq:temporal_dynamic}
    \mathbf{\Delta}_i^{(t)} = \Phi_{\rm map}(|\mathcal{T}_i^{(t-1)}-\mathcal{T}_i^{(t-2)}|)\in\mathbb{R}^{H\times W},
\end{equation}
where $\mathcal{T}_i^{(t)}$ is the object trajectory state generated in the Kalman filter, $|\cdot|$ denotes the L1-norm, which is used to quantify the overall difference in object states across various dimensions, and the mapping function $\Phi_{\rm map}(\cdot)$ assigns these differences to their respective coordinates in the BEV map.

\textbf{Temporal sampling.} The sampler outputs a binary temporal sampling matrix to determine whether each spatial region at current timestamp is sampled into the to-be-sent messages. To implement $\Phi_{\rm sample}(\cdot)$, we consider two aspects: i) initialize a uniform sampling interval for the scene. This ensures the unpredictable emergent object dynamics and the accumulated object dynamic estimation error can be captured, and ii) select additional timestamps when the dynamic level is above a predefined threshold. This helps ensure that significant changes in the object's state are captured. Let the initialize sampling interval be $\Delta T$, the dynamic threshold be $\sigma_t$, then, for each timestamp $t$, the sampling matrix $\mathbf{T}_{i}^{(t)}$ is given by
\begin{equation}
\vspace{-1mm}
\label{eq:temporal_compressor}
\mathbf{T}_{i}^{(t)}=\left\{
\begin{array}{lll}
\mathbf{1} \in \mathbb{R}^{H\times W},  &\quad t\%\Delta T=0;\\
\mathbf{\Delta}_i^{(t)}>\delta_t\in \mathbb{R}^{H\times W}, &\quad t\%\Delta T\neq0;
\vspace{-1mm}
\end{array} \right.
\end{equation}
where $\%$ denotes the modulo operation which performs the division and returns the resulting remainder, and $\mathbf{1}$ denotes the one-padded matrix, indicating selected. Note that by activating the communication system at the essential sampled timestamps, the overall communication cost can be effectively reduced. 

Referring to the strategy outlined in Section~\ref{sec:strategy}, supporters use pragmatic message selection to select the demanded information for each collaborator's downstream perceptual task. The optimized spatial and temporal selection matrices $\mathbf{S}_i^{(t)}$ and $\mathbf{T}_i^{(t)}$ specify $i$-th supporter's available perceptually critical information. Our cross-temporal collaboration strategy enables inference of a collaborator's needs based on their prior information received in latest communication round. More specifically, the $j$-th collaborator's request $\overline{\mathbf{S}}_{j}^{(t-1)}$ is defined as the reverse of its available information $\mathbf{S}_{j}^{(t-1)}$, this is, $\overline{\mathbf{S}}_{j}^{(t-1)}=1-\mathbf{S}_{j}^{(t-1)}$. The resulting sparse feature map $\mathcal{Z}_{i\rightarrow j}^{(t)}$ sent from agent $i$ to agent $j$ is given by
\begin{align}
\mathcal{Z}_{i\rightarrow j}^{(t)}&=\mathcal{F}_i^{(t)}\odot\mathbf{S}_{i}^{(t)}\odot\mathbf{T}_{i}^{(t)}\odot \overline{\mathbf{S}}_{j}^{(t-1)} \in \mathbb{R}^{H\times W \times C}.\label{subeq:func_selection}
\end{align}
This equation achieves the pragmatic message selection in Equation~\eqref{eq:message}. This selected feature map provides spatially and temporally sparse yet perceptually beneficial information, aligning with what supporters can provide and collaborators need. Note that, i) the sparsity of $\mathcal{Z}_{i\rightarrow j}^{(t)}$ is governed by binary matrices that dynamically manage the communication budget in different spatial-temporal regions based on their importance, accommodating various communication scenarios; and ii) only selected non-zero regions are included in messages, ensuring low communication costs.

\textbf{Advantages of pragmatic message selection.}
Compared to existing methods~\cite{LiuWhen2com:CVPR20,LiuWho2com:ICRA20}, our pragmatic message selection offers three key benefits: i) they indistinguishably transmit all the available information of the supporter, we tailor communication to meet the specific needs of collaborators; ii) they isolate communication at each timestamp, jointly consider the entire collaboration duration, reducing the temporal redundancy; and iii) they adopt an all-or-nothing strategy that handles the entire feature map uniformly, we use a finer-grained selection to prioritize the most beneficial spatial-temporal regions within available communication budgets, thereby promoting perception-communication trade-off across the entire communication bandwidth range.



\begin{figure}[!t]
    \centering
    \includegraphics[width=1.0\linewidth]{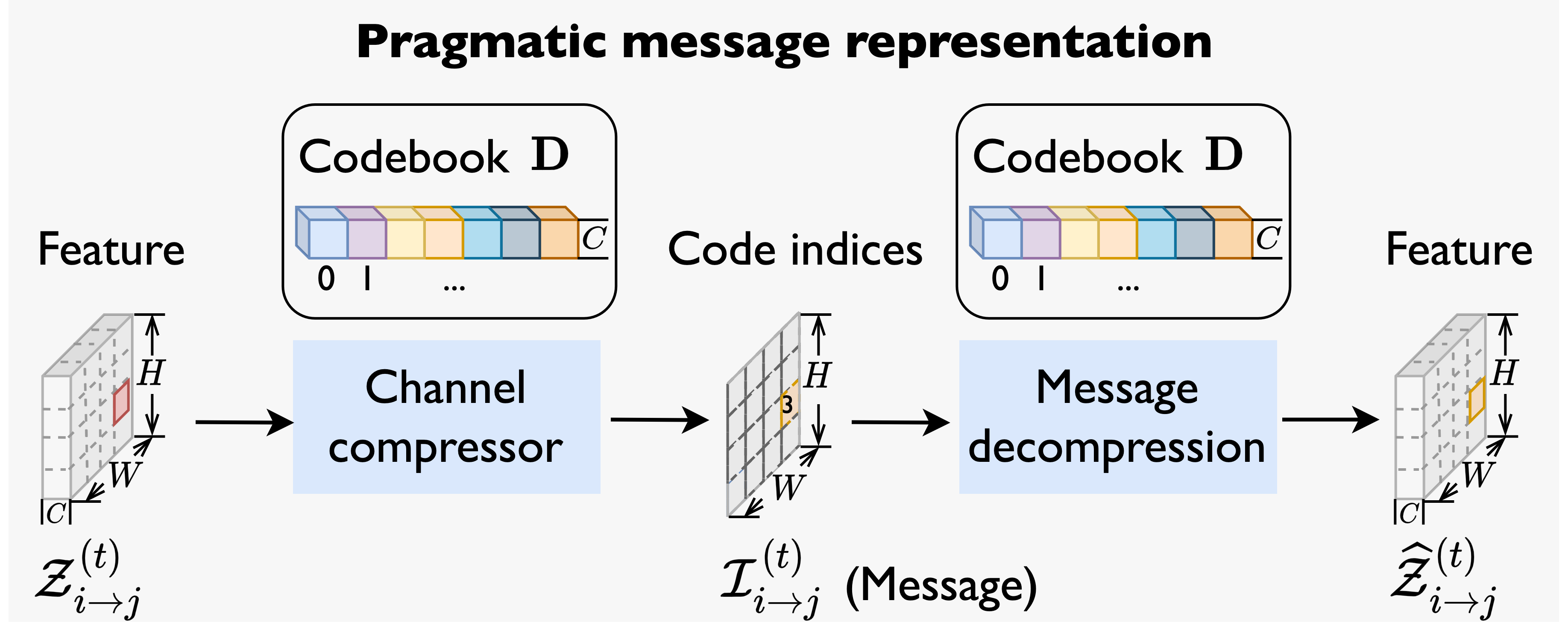}
    \vspace{-6mm}
    \caption{Overview of the channel compressor. The channel compressor transforms the dense feature representation into the lightweight code index representation.}
    \vspace{-6mm}
    \label{fig:codebook}
\end{figure}

\vspace{-1mm}
\subsubsection{Pragmatic message representation: Channel compressor}
\label{sec4:what2comm}

The pragmatic message selection leverages spatial and temporal compressors to reduce communication costs by selecting the sparse yet task-critical spatial temporal features from the complete feature map. Most previous works directly transmit these high-dimensional feature vectors, still incurring substantial communication expenses.
To address this issue, pragmatic message representation leverages the channel compressor to achieve a novel code message representation, reducing communication costs along the channel dimension. The core idea is to generate a pragmatical approximation of the high-dimensional feature vector with the most relevant code from a task-driven codebook; as a result, only integer code indices need to be transmitted, rather than the complete feature vectors composed of floating-point numbers.

Specifically, the channel compressor obtains the code representation with two steps: i) codebook learning which obtains the locally optimal codebook $\mathbf{D}\in\mathbb{R}^{n_L\times n_C}$ in~\eqref{subeq:opt_message}; and ii) pragmatic approximation which obtains the code index representation with the function $\Phi_{\mathbf{D}}(\cdot)$ in~\eqref{subeq:opt_message}.

\textbf{Codebook learning.} Analogous to a language dictionary used by humans, our task-driven codebook is shared among all agents to standardize their communication for achieving the detection task. This codebook consists of a set of codes, which are learned to pragmatically approximate possible perceptual features present in the training dataset. Here the pragmatic approximation refers to each code serving as a lossy approximation of a feature vector, while retaining necessary information for the downstream perception task within that vector. Let $\digamma=\{\mathcal{F}^{(i,\tau,s)}\}_{i=1,\tau=1,s=1}^{N,T,S}$ be the collective set of extracted BEV feature maps of all $N$ agents over $T$ time duration across all $S$ training scenes. Let $\mathbf{D} =   \begin{bmatrix} \mathbf{d}_1, \mathbf{d}_2, \cdots, \mathbf{d}_{n_L} \end{bmatrix}
\in\mathbb{R}^{C \times n_L}$ be the codebook, where $\mathbf{D}_{[\ell]} = \mathbf{d}_{\ell} \in \mathbb{R}^C$ is the $\ell$th code and $n_L$ is the code amount.

The task-driven codebook is learned through feature approximation at each spatial location; that is,
\begin{eqnarray}
\label{eq:codebook_learning}
\mathbf{D}^*= \arg\min_{\mathbf{D}} \sum_{\mathcal{F}\in\digamma}
\sum_{h,w} \min_{{\ell}} \left(  \Psi( \mathbf{D}_{[\ell]} )  + \left\| \mathcal{F}_{[h,w]} - \mathbf{D}_{[\ell]} \right\|_2^2 \right),
\end{eqnarray}
\vspace{-1mm}
where $\Psi(\cdot)$ denotes the resulting detection performance achieved by substituting $\mathbf{D}_{[\ell]}$ for $\mathcal{F}_{[h,w]}$. The first term pertains to the requirements of the downstream detection task and the second term reflects the reconstruction error between the original feature vector and the code. This approximation is lossy for reconstruction while lossless for the perceptual task, enabling the reduction of communication cost without sacrificing perceptual capacity.

\textbf{Code index representation.} 
Based on the shared codebook $\mathbf{D}$, each agent can substitute the selected sparse feature map $\mathcal{Z}_{i\rightarrow j}^{(t)}$ by a series of code indices $\mathcal{I}_{i\rightarrow j}^{(t)}$. For each BEV location $(h,w)$, the code index is obtained as,
\vspace{-1mm}
\begin{eqnarray}
\label{eq:code_index}
{(\mathcal{I}_{i\rightarrow j}^{(t)})}_{[h,w]} = \arg \min_{\ell} \left\| {(\mathcal{Z}^{(t)}_{i\rightarrow j})}_{[h,w]} - \mathbf{D}_{[\ell]}\right\|_2^2.
\end{eqnarray}
The codebook offers versatility in its configuration by adjusting both the codebook size $n_L$ and the quantity of codes $n_R$ used for representing the input vector. Equation~\eqref{eq:code_index} demonstrates a specific instance where $n_R=1$, chosen for simplicity in notation. When $n_R$ is larger, the representation involves a combination of multiple codes.

Overall, the final message sent from the $i$th agent to the $j$th agent is $\mathcal{P}_{i\rightarrow j}^{(t)}=\mathcal{I}_{i\rightarrow j}^{(t)}$, conveying the demanded complementary information with compact code indices. Only code indices of the selected feature vectors are packed, and they can be recovered at the receiver. 


\textbf{Advantages of pragmatic message representation.} Compared to feature messages~\cite{XuOPV2V:ICRA22,LiuWhen2com:CVPR20,HuWhere2comm:NeurIPS22,Hu2023CollaborationHC,LiLearning:NeurIPS21}, our code message representation offers i) efficiency for sending lightweight code indices; ii) adaptability to various communication resources via adjusting code configurations (smaller for efficiency, larger for superior performance), and iii) extensibility by offering a shared standardized representation. New heterogeneous agents can easily join the collaboration by adding its perceptual feature basis to the codebook.

\vspace{-1mm}
\subsubsection{Pragmatic collaborator selection: Message exchange}
\label{sec4:exchange}
Previous message compressors achieve pragmatic message selection and representation, resulting in sparse yet critical messages. Here, pragmatic collaborator selection targets to identify the most beneficial collaborators for exchanging these compact messages. The core idea is to establish connections only when beneficial information is available, this is, the sender can provide and simultaneously the receiver demands. 
Then it can effectively reduce unnecessary connections and curb the quadratically increased communication costs that accompany the collaborative agents increase.

Specifically, we achieve pragmatic collaborator selection by constructing a sparse communication graph, resulting in optimized adjacent matrix $\mathbf{A}$ in Equation~\eqref{subeq:opt_message}.
For each timestamp, we examine if the communication between agent $i$ and agent $j$ is necessary based on the maximum value of the binary selection matrix $\mathbf{S}_i^{(t)}\odot\mathbf{T}_i^{(t)}\odot\overline{\mathbf{S}}_j^{(t-1)}$, i.e. if there is at least one patch is activated, then we regard the connection is necessary. Formally, the $(i,j)$th element of the adjacency matrix $\mathbf{A}^{(t)}\in \{0, 1\}^{N\times N}$ of the communication graph at the $t$th communication round reflecting message passing from the $i$th agent to the $j$th agent is 
\begin{equation}
\setlength\abovedisplayskip{1pt}
\setlength\belowdisplayskip{1pt}
\mathbf{A}_{i,j}^{(t)}=\underset{\footnotesize{ h \in \{0,1,..,H-1\},w \in \{0,1,...,W-1\}}}{\rm max}~ \left( \mathbf{S}_i^{(t)}\odot\mathbf{T}_i^{(t)}\odot\overline{\mathbf{S}}_j^{(t-1)} \right)_{h,w} ,\label{subeq:collaborator_func}
\end{equation}
where $h, w$ index the spatial area. The sender's available support is represented with $\mathbf{S}_i^{(t)}\odot\mathbf{T}_i^{(t)}$, and the receiver's demand is represented with $\overline{\mathbf{S}}_j^{(t)}$. If this beneficial information is present in the messages, the collaborator is chosen.
Overall, the message sent from the $i$th agent to the $j$th agent at $t$-th timestamp is $\mathcal{P}_{i\rightarrow j}^{(t)}=\mathbf{A}_{i,j}^{(t)}\mathcal{I}_{i\rightarrow j}^{(t)}$. 

 
\textbf{Advantages of pragmatic collaborator selection.} 
Compared to existing works~\cite{XuOPV2V:ICRA22,XuV2V4Real:CVPR23,WangV2vnet:ECCV20,LiLearning:NeurIPS21} that leverage a fully connected communication graph, we prune the unnecessary connections, resulting a sparse graph, promoting efficiency. Compared to existing works~\cite{LiuWho2com:ICRA20,LiuWhen2com:CVPR20} that use attention to select partners with similar global features, we explicitly select the collaborators whose demanded information is available in the messages, fostering mutually beneficial collaboration and enhancing interpretability.

\vspace{-2mm}
\subsection{Pragmatic collaboration: Message utilization}
\label{sec4:utilization}

Referring to the analysis in Section~\ref{sec:formulation}, message utilization optimizes the second sub-optimization~\eqref{subeq:perfromance_subobjective}, maximizing the perception performance given the obtained pragmatic messages in Section~\ref{sec4:determination}. 
Following our design rationale of collaborating over the entire duration, message utilization incorporates all available pragmatic messages to complement single-agent observations for a more thorough perception. The intuition for integrating the full history of pragmatic messages, both past and present, is twofold: i) to reconstruct a complete perceptual sequence from temporally sampled pragmatic messages, and ii) to offer temporal complementarity, since targets obscured in the current timestamp might have been visible in previous ones.

To achieve this, message utilization generates the collaborative feature with two steps: i) message decoding, which involves the decompression function $\Phi_{\mathbf{D}}^{-1}(\cdot)$ to decode perceptual features from current code-index-based pragmatic messages, and the prediction function $\Phi_{\rm pred}(\cdot)$ to estimate complementary perceptual features at the current state using historical collaborative data; and ii) message fusion, which aggregates all decoded perceptual information to enhance individual perceptual features.
We avoid reprocessing past pragmatic messages by using accumulated collaborative features, assuming each timestamp's collaborative feature optimally aggregates all available data to that point.
Then, the collaborative feature $\mathcal{H}_i^{(t)}$ at timestamp $t$ is given by
\begin{subequations}
\setlength\belowdisplayskip{1pt}
\begin{align}
\mathcal{H}_i^{(t)}&=\Phi_{\text{fusion}}\left(\mathcal{F}_i^{(t)},\{\Phi_{\mathbf{D}}^{-1}\left(\mathcal{P}_{j\rightarrow i}^{(t)}\right)\}_{j=1}^N, \Phi_{\rm pred}\left(\{\mathcal{H}_i^{(k)}\}_{k=1}^{t-1}\right)\right).
\end{align}    
\end{subequations}
Upon generation, the collaborative feature replaces the original single-agent feature $\mathcal{F}_i^{(t)}$ and is then decoded into detections $\widetilde{\mathcal{O}}_i^{(t)}$ and tracks $\widetilde{\mathcal{T}}_i^{(t)}$. 
The decompression and prediction functions, $\Phi_{\mathbf{D}}^{-1}(\cdot)$ and $\Phi_{\rm pred}(\cdot)$, are achieved in Section~\ref{sec4:decompression}, and the fusion function $\Phi_{\rm fusion}(\cdot,\cdot,\cdot)$ is achieved in Section~\ref{sec4:fusion}.

\vspace{-1mm}
\subsubsection{Message decoding}
\label{sec4:decompression}
Message decoding targets to decode the complementary information from the collaborative message sequences, including two parts: i) message decompression decodes current pragmatic messages, and ii) message prediction estimates the current state of historical collaborative data.

\textbf{Message decompression.} Message decompression reconstructs the supportive features based on the currently received code indices and the shared codebook. Given the received message $\mathcal{P}_{j\rightarrow i}^{(t)}={\mathcal{I}}_{j\rightarrow i}^{(t)}$, the decoded feature map's $\widehat{\mathcal{Z}}_{j\rightarrow i}^{(t)}\in\mathbb{R}^{H\times W\times C}$ element located at $(h,w)$ is given by
\begin{equation}
\setlength\abovedisplayskip{1pt}
\setlength\belowdisplayskip{1pt}
    (\widehat{\mathcal{Z}}_{j\rightarrow i}^{(t)})_{[h,w]}=\mathbf{D}_{[(\mathcal{I}_{i\rightarrow j}^{(t)})_{[h,w]}]}.
\end{equation}
Subsequently, message fusion aggregates these decoded feature maps to augment individual features.


\begin{figure}[!t]
    \centering
    \includegraphics[width=1.0\linewidth]{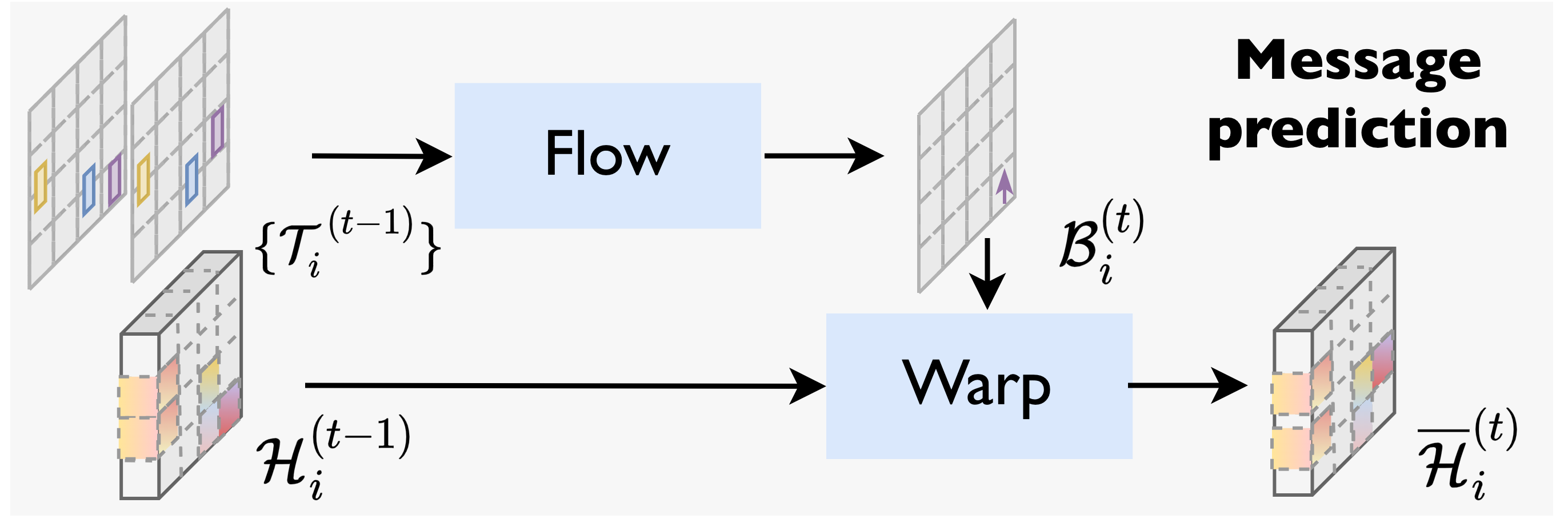}
    \vspace{-6mm}
    \caption{Overview of the prediction module. It aligns the historical feature map with the current timestamp, using a warp function and the predicted displacements obtained in the flow estimation function.}
    \label{fig:prediction}
    \vspace{-6mm}
\end{figure}

\textbf{Message prediction.} Message prediction estimates the current perceptual state of historical perceptual feature sequences. It allows for reconstructing the complete perceptual sequences given the temporally sampled pragmatic information in Section~\ref{sec4:when2comm}. 
Message prediction is accomplished through two steps: i) creating the Bird's Eye View (BEV) flow map $\mathcal{B}_{i}^{(t)}\in\mathbb{R}^{H\times W \times 2}$, which captures the movement of perceptual targets, and ii) estimating the current perceptual state $\overline{\mathcal{H}}_{i}^{(t)}\in\mathbb{R}^{H\times W \times C}$ by compensating historical perceptual features based on the calculated flow.



For Step i), we generate the BEV flow map with two steps: 1) estimate motion vectors $\Delta\mathcal{T}_{i}^{(t)}$ for each object; and 2) map these motion vectors to the BEV feature coordinate system, resulting in BEV flow map $\mathcal{B}_{i}^{(t)}$. This is implemented by
\begin{subequations}
\begin{align}
\label{eq:motion}
\Delta\mathcal{T}_{i}^{(t)}&= (\overline{\mathcal{T}}_{i}^{(t)}-\mathcal{T}_{i}^{(t-1)})_{xy}\in\mathbb{R}^{\overline{M}_i^{(t-1)}\times 2},\\
\label{eq:flow_map}
\mathcal{B}_{i}^{(t)}&= \Phi_{\rm map}\left(\Delta\mathcal{T}_{i}^{(t)}\right)\in\mathbb{R}^{H\times W \times 2}.
\end{align}    
\end{subequations}
Here, we reuse the Kalman filter (Section~\ref{sec4:tracker}) to model the object dynamics and predict the trajectory at current timestamp. Then the motion vectors are calculated by the difference between the predicted trajectory $\overline{\mathcal{T}}_{i}^{(t)}$ and the latest trajectory $\mathcal{T}_{i}^{(t-1)}$, achieving 1) in~\eqref{eq:motion}.
Subsequently, we use the same as the mapping function $\Phi_{\rm map}(\cdot)$ defined in Equation~\eqref{subeq:temporal_dynamic} to assign the input vectors to their respective coordinates in the BEV map, achieving 2) in~\eqref{eq:flow_map}.

For Step ii), we update the latest feature map to reflect the current state by realigning features according to their estimated positions, derived from motion. Utilizing the predicted BEV flow map and the historical feature $\mathcal{H}_{i}^{(t-1)}$, we generate the current feature map $\overline{\mathcal{H}}_{i}^{(t)}\in\mathbb{R}^{H\times W \times C}$. This is done by mapping each element at coordinates $(h \in \{0,1,..,H-1\},w \in \{0,1,...,W-1\})$ as follows
\vspace{-1mm}
\begin{equation}
\vspace{-1mm}
{(\overline{\mathcal{H}}_{i}^{(t)})}_{[h+{(\mathcal{B}_{i}^{(t)})}_{[h,w,0]},w+{(\mathcal{B}_{i}^{(t)})}_{[h,w,1]}]} = {(\mathcal{H}_{i}^{(t-1)})}_{[h,w]}.
\end{equation}
Once this updated feature map is generated, it is then output to the fusion module to enhance the single-agent features. Note that, this updated feature map offers two key advantages: i) it provides temporal complementarity, and ii) it provides robustness, as intermediate features enable the flexibility of filtering out inaccurate perceptual information during aggregation and subsequent decoding phases.

\vspace{-1mm}
\subsubsection{Message fusion}
\label{sec4:fusion}

Message fusion augments the single-agent feature by aggregating the decoded feature maps. We implement this with simple but effective non-parametric point-wise maximum fusion.
Specifically, for the $i$th agent, given the decompressed feature maps $\overline{\mathcal{Z}}_{j\rightarrow i}^{(t)}$ from the collaborative neighbors $\mathcal{N}_i^{(t)}$ or the estimated feature map $\overline{\mathcal{H}}_{i}^{(t)}$. We also include the ego feature map in fusion. The fused BEV feature is obtained as
\vspace{-1mm}
\begin{equation}\label{eq:fusion}
\vspace{-1mm}
\mathcal{H}_i^{(t)}=\underset{j\in\mathcal{N}_i}{\rm max}(\mathcal{F}_{i}^{(t)},{\widehat{\mathcal{Z}}}^{(t)}_{j\rightarrow i},\overline{\mathcal{H}}_{i}^{(t)})\in\mathbb{R}^{H\times W\times D},
\end{equation}
where ${\rm max}(\cdot)$ maximizes the corresponding features from multiple agents at each individual spatial location. Note that attention fusion is not permutation invariant, as attention weights vary with the ordering of key and query. Here we simply use the max operator to avoid this permutation variant issue. Note that the enhanced feature $\mathcal{H}_i^{(t)}$ substitutes the single-agent feature $\mathcal{F}_i^{(t)}$ and sequentially outputs to the decoder and the tracker to generate the upgraded detection $\widetilde{\mathcal{O}}_{i}^{(t)}$ and tracking trajectories $\widetilde{\mathcal{T}}_{i}^{(t)}$. 

\vspace{-2mm}
\subsection{Loss functions}
\label{sec4:loss}

To train the overall system, we supervise three tasks: spatial confidence map generation, object detection, and codebook learning. The spatial confidence map generator reuses the parameters of the detection decoder. The overall loss is defined as
\begin{equation}
\setlength\abovedisplayskip{2pt}
\setlength\belowdisplayskip{1pt}
L = \sum_{t=1}^T\sum_i^{N} L_{\text{det}} \left(\widetilde{\mathcal{O}}_i^{(t)},\widehat{\mathcal{O}}_i^{(t)} \right)+\left\|{\overline{\mathcal{F}}}_{i}^{(t)},{\mathcal{F}}_{i}^{(t)}\right\|_2^2, 
\end{equation}
where $L_{\text{det}}(\cdot)$ denotes the detection loss~\cite{ZhouObjects:Arxiv2019}, $\widetilde{\mathcal{O}}_i^{(t)}$ and $\widehat{\mathcal{O}}_i^{(t)}$ represents the ground-truth and predicted objects, and ${\mathcal{F}}_{i}$ and ${\overline{\mathcal{F}}}_{i}$ denote the $i$-th agent's original feature map and the one approximated by codes. During the optimization, the network parameters and the codebook are updated simultaneously. This combined supervision encourages the codebook to preserve critical perceptual information, enabling the pragmatic code representation to be lossless for downstream perception tasks. This approach enhances communication efficiency without sacrificing perception performance.



\textbf{Advantages of pragmatic collaboration implementation.} 
This neural-network-based pragmatic collaboration implementation actualizes the pragmatic communication strategy outlined in Section~\ref{sec:formulation} and addresses the two sub-optimization problems.
First, it addresses the first sub-optimization, message determination, through message compression which selects and represents pragmatic information from spatial, temporal, and channel aspects, and message exchange, which selects the pragmatic beneficial collaborators. By optimizing for sparsity across time, space, and the communication graph, as well as refining the granularity of the representation, this implementation ensures efficient pragmatic messages under diverse communication conditions.
Second, it addresses the second sub-optimization, message utilization, by employing message decoding to extract valuable information from both historical and recent pragmatic messages, and message fusion to integrate this information and enhance individual perception capabilities. By leveraging the beneficial information from spatial and temporal pragmatic messages, this implementation improves perception performance.

Furthermore, our pragmatic collaboration implementation not only has advantages in performance-communication trade-off, but also in compatibility to address practical issues. 
For the latency issue, the prediction module within the message decoding process can temporally compensate for delayed messages, synchronizing them with the current timestamp for fusion.
For the heterogeneous issues, the shared codebook establishes a common feature space, facilitating the alignment of diverse data sources. This is akin to a form of linguistic translation in human communication, where each agent converts their local language into a universally understood language.



\vspace{-2mm}
\section{Experiments and analysis}
\label{sec:experiment}
In this section, we evaluate the proposed~\texttt{PragComm}. First, we introduce the datasets and model settings in detail. Second, we present the performance comparisons between~\texttt{PragComm} and other state-of-the-art methods. Third, we show the ablation studies and visualization results.

\begin{figure*}[!ht]
  \centering
  \includegraphics[width=1.0\linewidth]{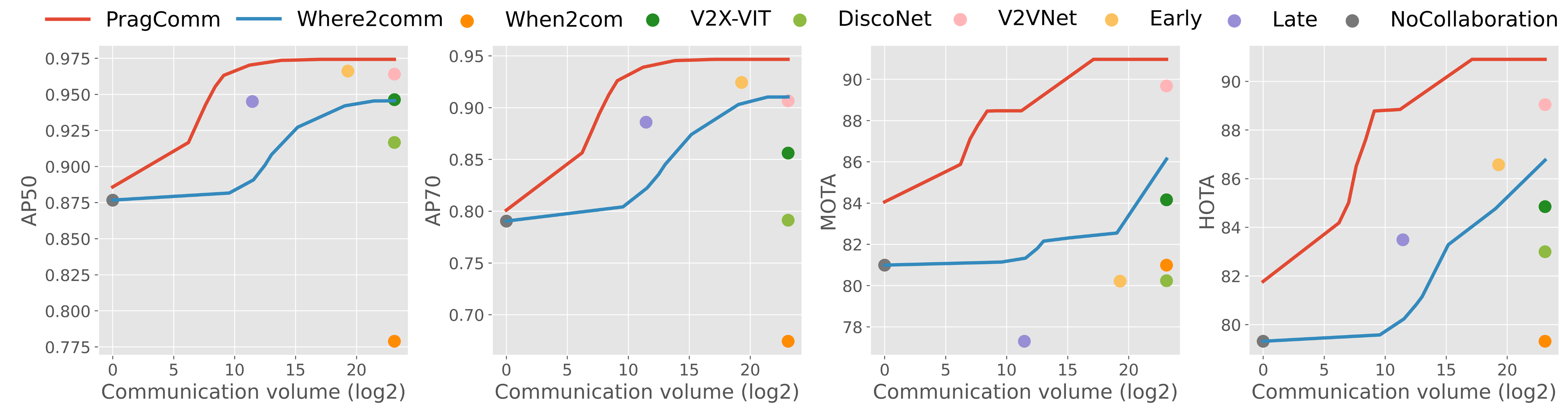}
  \vspace{-8mm}
  \caption{Comparison with SOTAs on OPV2V. PragComm steadily improves 3D detection and tracking performance as the communication resource grows, and outperforms previous SOTA methods across dynamic communication conditions.}
  \vspace{-6mm}
  \label{Fig:OPV2V_SOTAs}
\end{figure*}

\vspace{-1mm}
\subsection{Datasets and Model Setting}

We conduct experiments on both collaborative 3D object detection and tracking tasks using three datasets, including the real-world dataset, V2V4Real~\cite{XuV2V4Real:CVPR23}, and simulation datasets, OPV2V~\cite{XuOPV2V:ICRA22} and V2X-SIM2.0~\cite{LiV2XSim:RAL22}.
Our experiments include both homogeneous scenarios, where agents have identical sensors, and heterogeneous scenarios, randomly featuring different sensors (LiDAR/camera). 
We also evaluate practical issues, including pose errors and communication latency.

\textbf{V2V4Real.} V2V4Real is a large real-world vehicle-to-vehicle collaborative perception dataset. It includes a total of $20$K frames of LiDAR point cloud captured by Velodyne VLP-32 sensor and $40$K frames of RGB images captured by two mono cameras (front and rear) with $240$K annotated 3D bounding boxes. The perception range is 280m$\times$80m. 

\textbf{OPV2V.} OPV2V is a vehicle-to-vehicle collaborative perception simulation dataset, co-simulated by OpenCDA~\cite{Xu2021OpenCDAAO} and Carla~\cite{carla}. It includes $12$K frames of 3D point clouds and RGB images with $230$K annotated 3D boxes. The perception range is 280m$\times$80m. 

\textbf{V2X-SIM2.0.} V2X-SIM2.0 is a vehicle-to-vehicle collaborative perception simulation dataset. Each scene contains a 20-second traffic flow at a certain intersection of three CARLA towns, and the multi-modality multiagent sensory streams are recorded at 5Hz. It includes $47.2$K samples with $10$K frames of 3D point clouds and RGB images. The perception range is 100m$\times$80m. 



\textbf{Model architecture.} For the camera and LiDAR inputs, we implement the feature extractor following CADDN~\cite{CaDDN} and PointPillar~\cite{PointPillar}. For the heterogeneous setup, agents are randomly assigned either LiDAR or camera, resulting in a balanced 1:1 ratio of agents across the different modalities. The input feature maps from multiple modalities are transformed to BEV with a resolution of $0.8$m/pixel and fused together to complement each other.


\textbf{Evaluation metrics.}
Following the collaborative perception methods~\cite{HuWhere2comm:NeurIPS22,XuOPV2V:ICRA22,LiV2XSim:RAL22}, the detection results are evaluated by Average Precision (AP) at Intersection-over-Union (IoU) thresholds of 0.50 and 0.70.
The tracking results are mainly evaluated by HOTA (Higher Order Tracking Accuracy)~\cite{Luiten2020HOTAAH} to evaluate our BEV tracking performance. HOTA can evaluate detection, association, and localization performance via a single unified metric. In addition, the classic multi-object tracking accuracy (MOTA) and multi-object tracking precision (MOTP)~\cite{Bernardin2008EvaluatingMO} are also employed. MOTA can measure detection errors and association errors. MOTP solely measures localization accuracy.

\textbf{Communication volume.} Following the collaborative perception methods~\cite{LiLearning:NeurIPS21,HuWhere2comm:NeurIPS22,HuCollaboration:CVPR23}, to align with the metric bit/byte, the communication volume counts the message size by byte in log scale with base $2$. Specifically, given each selected feature vector, for feature representation, its bandwidth is calculated as ${\rm log}_2(C\times 32/8)$. Here, $32$ represents the float32 data type, and $8$ converts bits to bytes. For code index representation, given codebook $\mathbf{D}\in\mathbb{R}^{C\times n_L}$, comprised of $n_L$ codes and each vector constructed using $n_R$ codes, the bandwidth of each selected feature vector is calculated as ${\rm log}_2({\rm log}_2(n_L)\times n_R /8)$. Here, ${\rm log}_2(n_L)$ signifies the data amount required to represent each code index integer, decided by the codebook size.


\textbf{Training details.} For camera-only input, the model is trained $100$ epoch with initial learning rate of $1$e-$3$, and decayed by 0.1 at epoch 80. For the LiDAR input, we train $120$ epochs with a learning rate of $1$e-$3$. For multi-modality input, the model is trained $120$ epoch with an initial learning rate of $1$e-$3$, and decayed by 0.1 at epoch 80.


\vspace{-1mm}
\subsection{Benchmark comparison}
\textbf{Baselines.} To validate the effectiveness of the proposed method, we compare it with a bunch of baselines including\\
\noindent$\bullet$ \texttt{single}: single-agent perception without collaboration; \\
\noindent$\bullet$ \texttt{late}: late fusion where agents directly exchange the perceived 3D boxes, which consumes little communication cost while also inferior perception performance as each individual perception output could be noisy and incomplete, causing unsatisfying fusion results;\\
\noindent$\bullet$ \texttt{early}: early fusion where agents exchange the raw measurements, e.g. 3D point clouds or images, which consumes large communication cost while superior perception performance as the messages conserve the complete information; \\
\noindent$\bullet$ \texttt{intermediate}: intermediate fusion where agents exchange the intermediate features, which are widely adopted in previous collaborative methods~\cite{LiuWho2com:ICRA20,LiuWhen2com:CVPR20,WangV2vnet:ECCV20,LiLearning:NeurIPS21,XuOPV2V:ICRA22,XuV2XViT:ECCV22,XuCoBEVT:CoRL22,HuWhere2comm:NeurIPS22,Hu2023CollaborationHC} as this approach can squeeze representative information into compact features and can balance perception performance and communication cost trade-off. We include previous SOTAs, When2com~\cite{LiuWhen2com:CVPR20}, V2VNet~\cite{WangV2vnet:ECCV20}, DiscoNet~\cite{LiLearning:NeurIPS21}, V2X-ViT~\cite{XuV2XViT:ECCV22}, Where2comm~\cite{HuWhere2comm:NeurIPS22}, and HMViT~\cite{XiangHMViT:ICCV23}.


\begin{figure*}[!ht]
  \centering
  \includegraphics[width=1.0\linewidth]{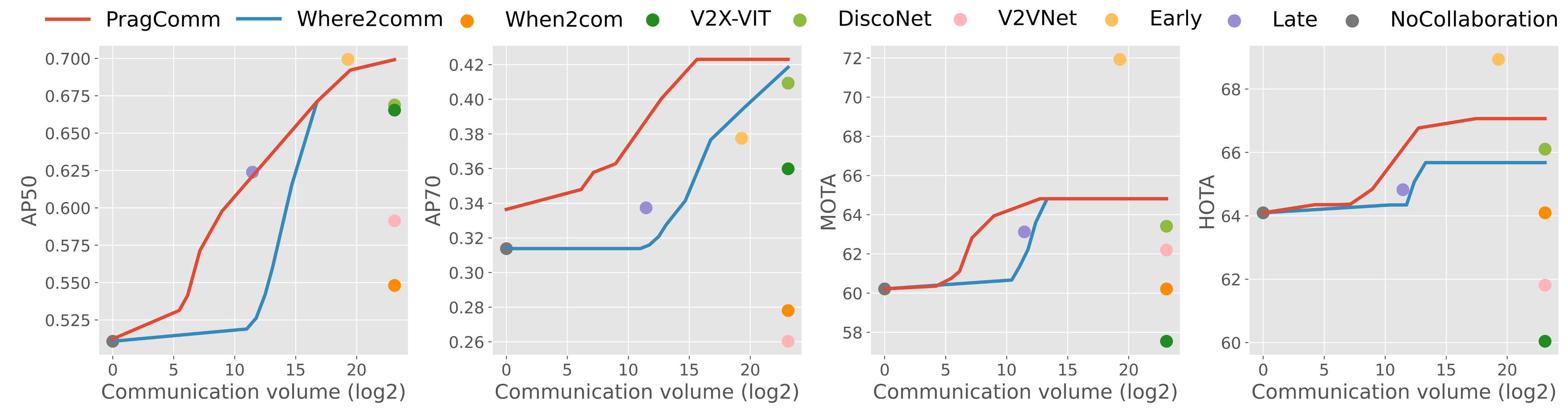}
  \vspace{-8mm}
  \caption{Comparison with SOTAs on V2V4Real. PragComm consistently outperforms previous SOTA methods across dynamic communication conditions on both 3D detection and tracking tasks.}
  \vspace{-5mm}
  \label{Fig:V2V4Real_SOTA}
\end{figure*}

\begin{figure*}[!ht]
  \centering
  \includegraphics[width=1.0\linewidth]{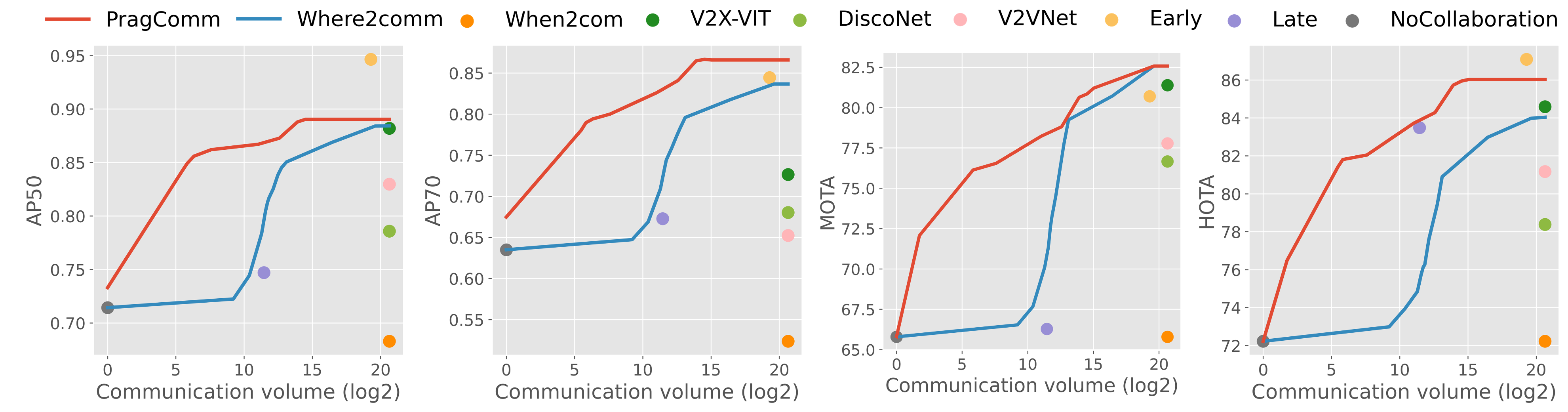}
  \vspace{-8mm}
  \caption{Comparison with SOTAs on V2X-SIM2.0. PragComm consistently outperforms previous SOTA methods across dynamic communication conditions on both 3D detection and tracking tasks.}
  \vspace{-5mm}
  \label{Fig:V2X-SIM2_SOTA}
\end{figure*}

\begin{figure*}[!ht]
  \centering
  \includegraphics[width=1.0\linewidth]{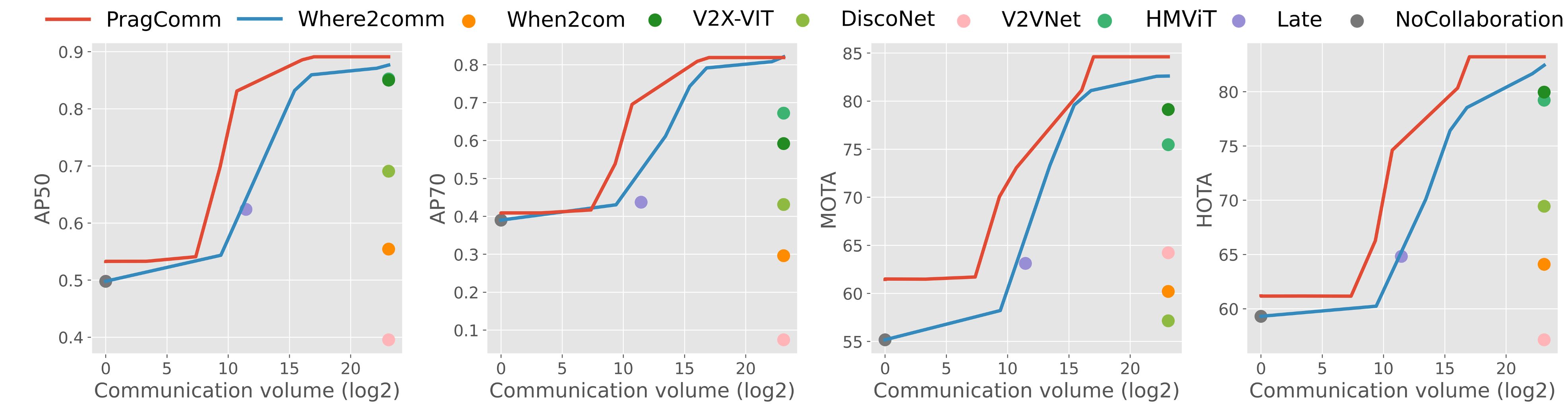}
  \vspace{-8mm}
  \caption{Comparison with SOTAs on heterogeneous setting on OPV2V. PragComm consistently outperforms previous SOTA methods across dynamic communication conditions on both 3D detection and tracking tasks.} 
  \vspace{-6mm}
  \label{Fig:OPV2VHete_SOTA}
\end{figure*}

\begin{figure*}[!ht]
  \centering
  \includegraphics[width=0.98\linewidth]{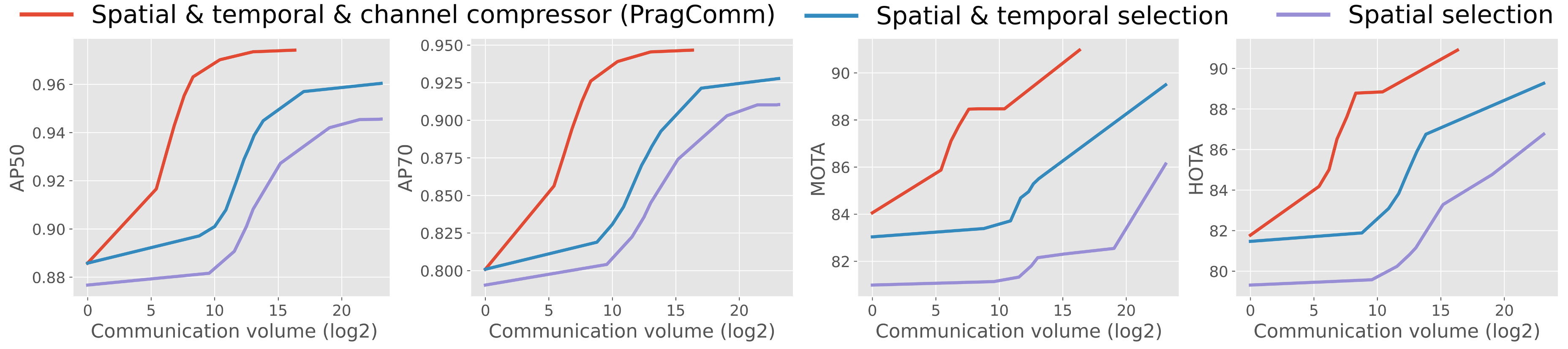}
  \vspace{-4mm}
  \caption{Effectiveness of the spatial, temporal, and channel compressor.}
  \vspace{-8mm}
  \label{Fig:ABL_STC}
\end{figure*}

\begin{figure}[!ht]
  \centering
  \begin{subfigure}{0.48\linewidth}
    \includegraphics[width=1.0\linewidth]{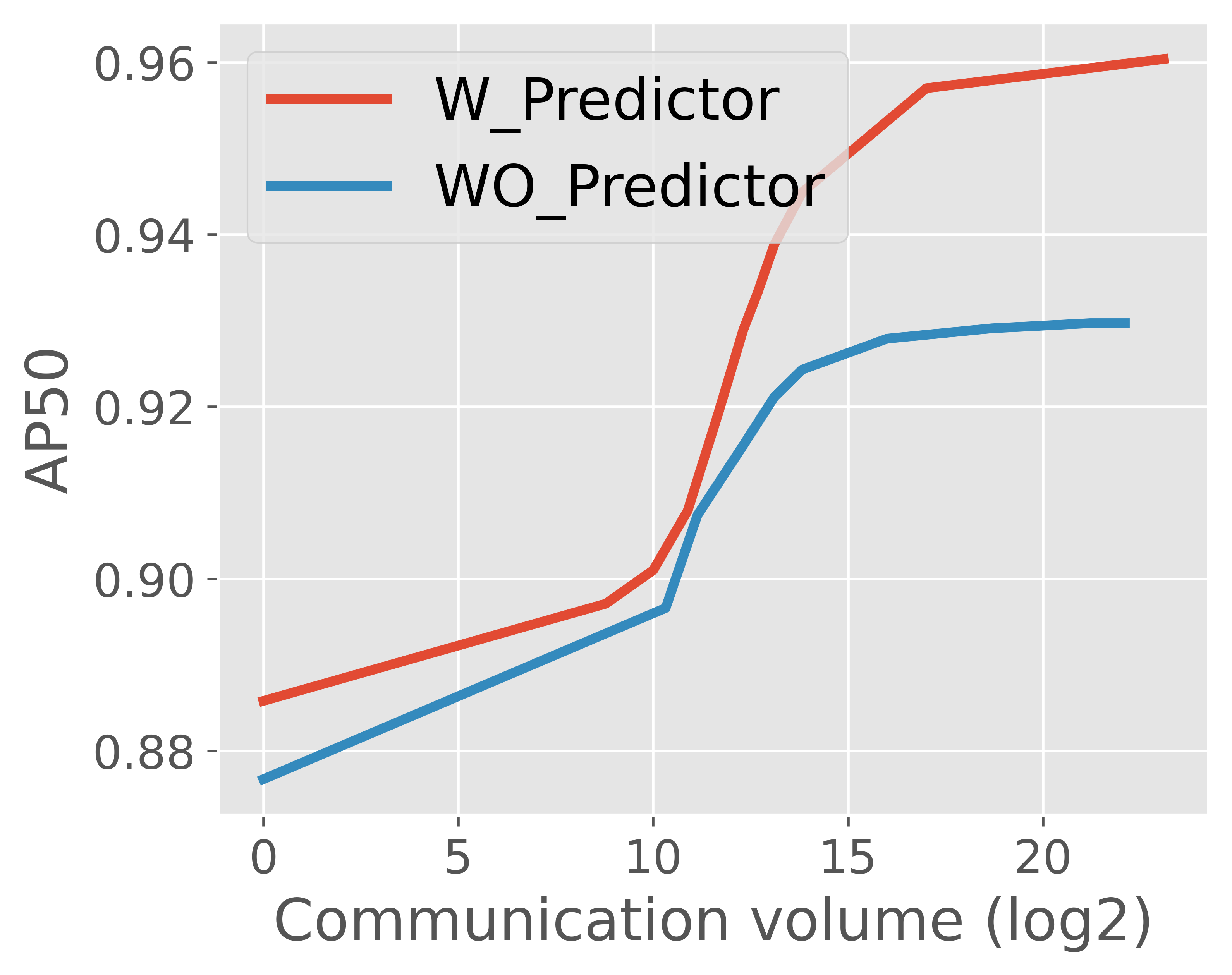}
  \end{subfigure}
  \begin{subfigure}{0.48\linewidth}
    \includegraphics[width=1.0\linewidth]{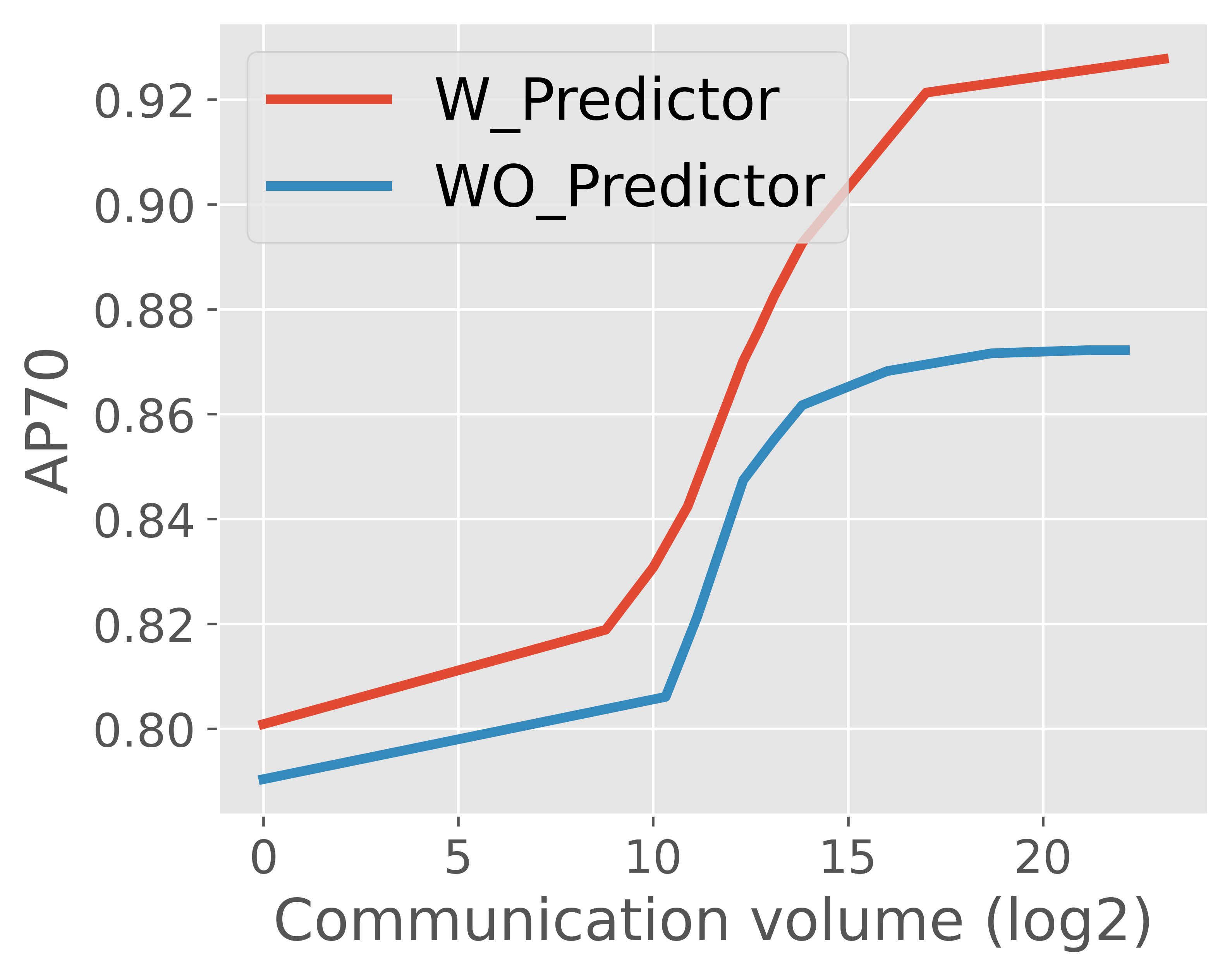}
  \end{subfigure}
  \vspace{-3mm}
  \caption{Effectiveness of predictor.}
  \vspace{-5mm}
  \label{Fig:ABL_Predictor}
\end{figure}

\begin{figure}[!ht]
  \centering
  \begin{subfigure}{0.48\linewidth}
    \includegraphics[width=1.0\linewidth]{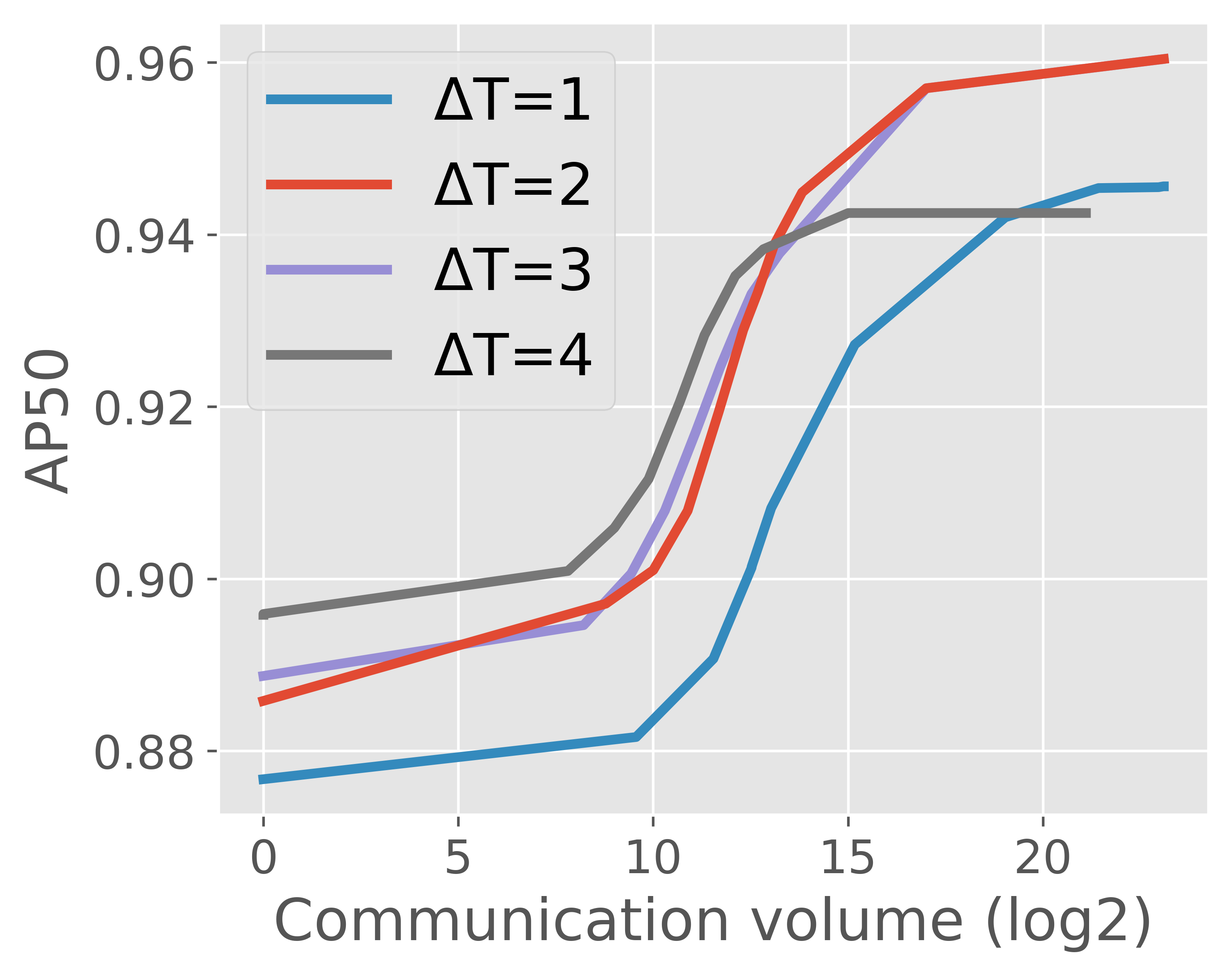}
  \end{subfigure}
  \begin{subfigure}{0.48\linewidth}
    \includegraphics[width=1.0\linewidth]{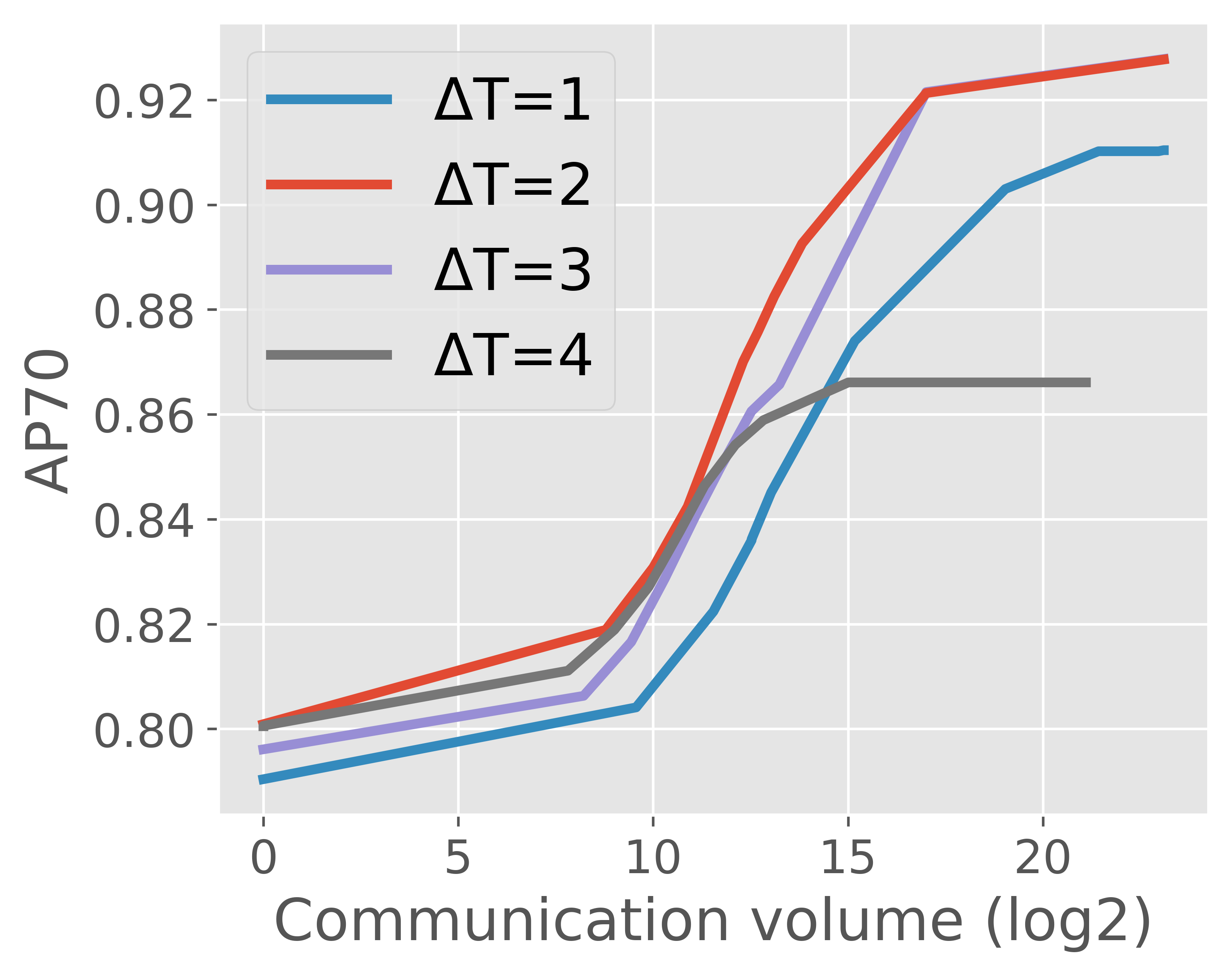}
  \end{subfigure}
  \vspace{-3mm}
  \caption{Effectiveness of temporal frequency.}
  \vspace{-7mm}
  \label{Fig:ABL_Temp}
\end{figure}

\textbf{Effectiveness of perception-communication trade-off.}
Fig.~\ref{Fig:OPV2V_SOTAs},~\ref{Fig:V2V4Real_SOTA},~\ref{Fig:V2X-SIM2_SOTA} and~\ref{Fig:OPV2VHete_SOTA} compares the proposed \texttt{PragComm} with the baselines in terms of the trade-off between detection performance (AP@IoU=0.50/0.7), tracking performance (MOTA/HOTA) and communication bandwidth on OPV2V, V2V4Real, and V2X-SIM2.0 under homogeneous settings and OPV2V under heterogeneous settings, respectively.
The red and blue curves come from a single~\texttt{PragComm} and~\texttt{Where2comm} model evaluated at varying bandwidths. We see that: i)~\texttt{PragComm} achieves a far-more superior perception-communication trade-off across all the communication bandwidth choices and various collaborative perception settings, including both 3D object detection and tracking tasks with lidar-only, and heterogeneous sensor setup; 
ii)~\texttt{PragComm} significantly outperforms previous state-of-the-arts (SOTAs) on both real-world (V2V4Real) and simulation scenarios, improves the SOTA performance by 7.7\% on V2V4Real, 25.81\% on OPV2V, 1.9\% on V2X-SIM2.0; 
iii)~\texttt{PragComm} achieves the same detection performance of previous SOTAs with extremely less communication volume: more than 32.7K times less on OPV2V, 55 times less on V2X-SIM2.0, 105 times less on V2V4Real, 32 times less on OPV2V heterogeneous;
and iv)~\texttt{PragComm} and~\texttt{Where2comm} can adjust to varying communication conditions while previous methods are limited to specific communication choices.

\vspace{-1mm}
\subsection{Ablation Studies}
\vspace{-1mm}
\textbf{Effectiveness of spatial, temporal, and channel compressors in message determination.} Fig.~\ref{Fig:ABL_STC} compares three variants of the message compressors on both detection and tracking tasks. We see that: i) applying spatial, temporal, and channel compressors substantially improves the trade-off between perception performance and communication cost. It supports our claim that the proposed compressors effectively eliminate orthogonal redundancy from three aspects, leading to more substantial efficiency improvement; and ii) applying temporal and channel compressors enhances the perception performance by 1.97\%/2.13\% and 4.10\%/4.31\% for detection/tracking on OPV2V with the same communication cost (16). The reasons are: a) the temporal compressor provides complementary information by fusing the historical features, especially for the currently occluded while historically visible objects; and b) the codebook-based channel compressor improves feature learning with the dictionary learning technology as it efficiently reduces the data noise by using a sparse representation, creating a more robust feature space.


\textbf{Effect of predictor in message utilization.}
Fig.~\ref{Fig:ABL_Predictor} compares two variants in message utilization: with/without message predictor.
Variant a) \texttt{WO\_Predictor} directly use the received collaborative features at the current timestamp;
In contrast, variant b) \texttt{W\_Predictor} forecasts historical collaborative features to the present state and merges this adjusted feature with the current timestamp's collaborative features.
We see that \texttt{W\_Predictor} significantly outperforms \texttt{WO\_Predictor} across all the communication bandwidth conditions.
This indicates that: i) our predictor effectively recovers the missed information at the unsampled timestamps given the temporally sampled pragmatic messages; and ii) the temporal compressor efficiently selects pragmatic messages that maintain all necessary information, enabling successful recovery.

\begin{figure}[!t]
  \centering
  \begin{subfigure}{0.48\linewidth}
    \includegraphics[width=1.0\linewidth]{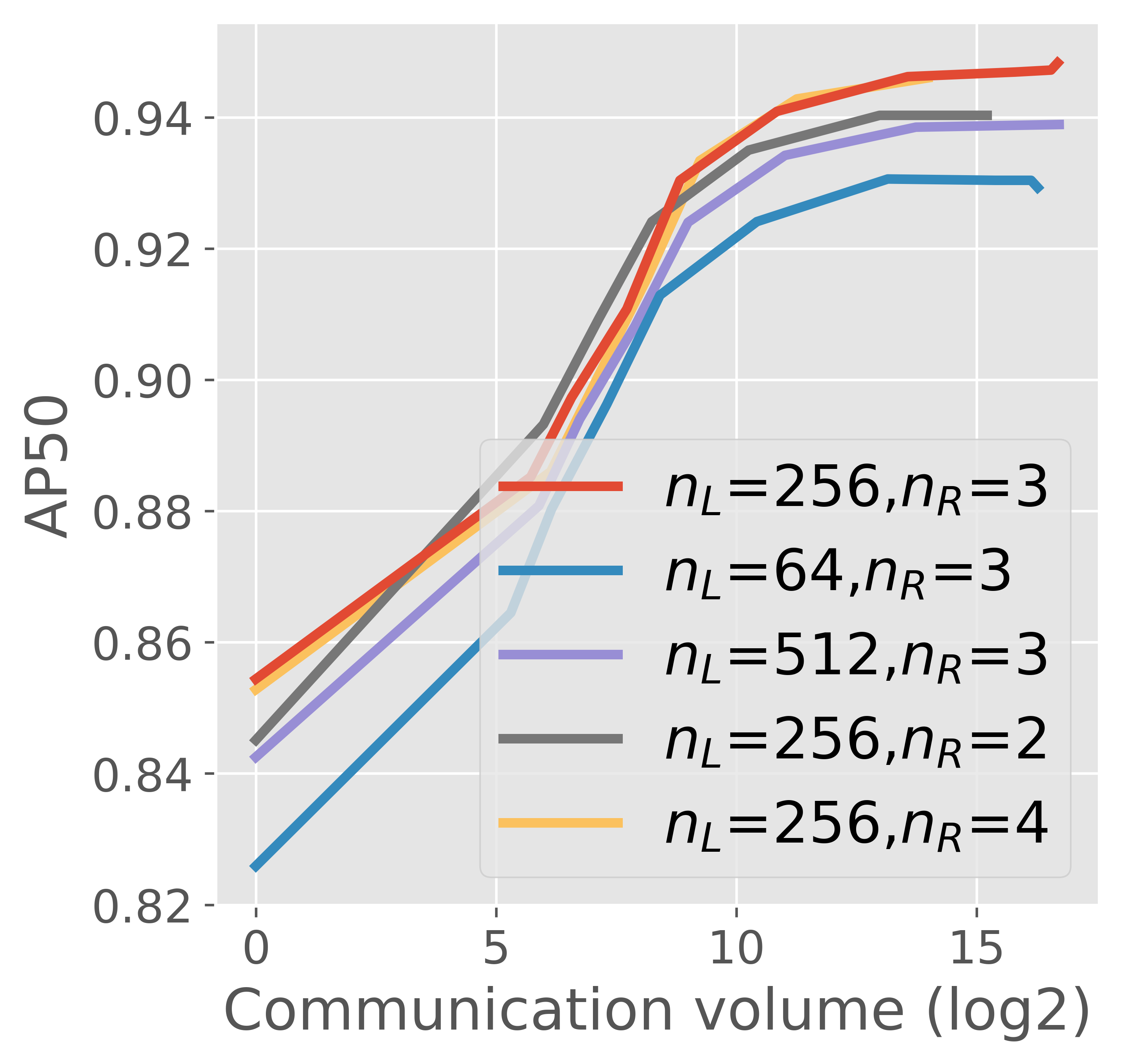}
  \end{subfigure}
  \begin{subfigure}{0.48\linewidth}
    \includegraphics[width=1.0\linewidth]{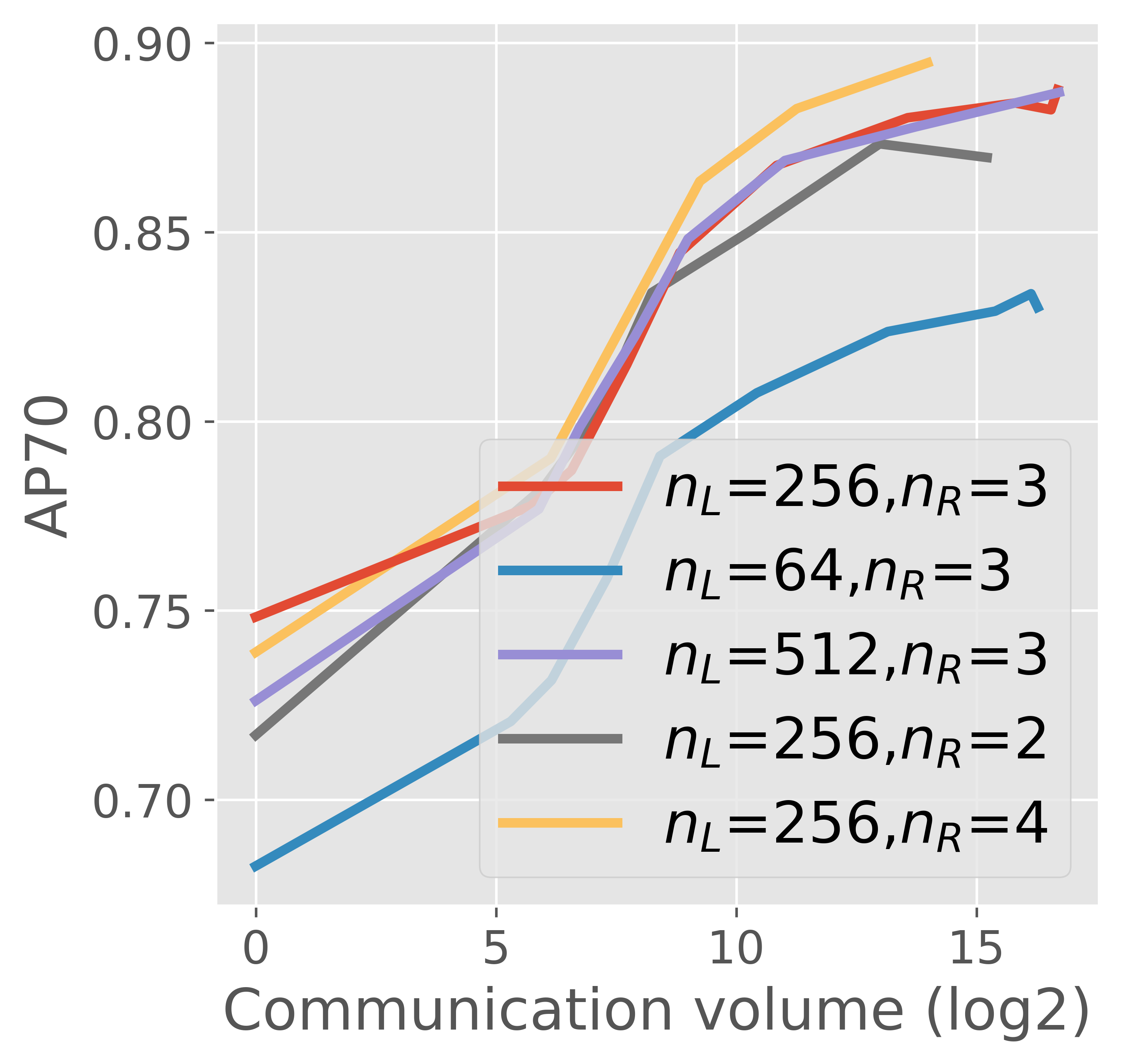}
  \end{subfigure}
  \vspace{-3mm}
  \caption{Effectiveness of codebook configurations, including codebook size $n_L$ and code quantity $n_R$.}
  \vspace{-6mm}
  \label{Fig:ABL_Codebook}
\end{figure}



\textbf{Effect of sampling interval in temporal compressor.}
Fig.~\ref{Fig:ABL_Temp} illustrates the comparison of four sampling intervals used in the temporal compressor, where $\Delta T=1$ denotes no temporal sampling and compensation. We see that: 
i) at an appropriate sampling frequency, the temporal compressor enhances the performance-communication trade-off, achieving an optimal balance between perception quality and communication efficiency; and
ii) the performance-communication trade-off initially improves and then deteriorates with increasing temporal sampling intervals. This aligns with the Nyquist sampling theorem, indicating that a sampling frequency lower than the object dynamics adversely affects the predictor and recovery of the object sequences.

\begin{figure*}[!t]
\vspace{-1mm}
  \centering
  \includegraphics[width=1.0\linewidth]{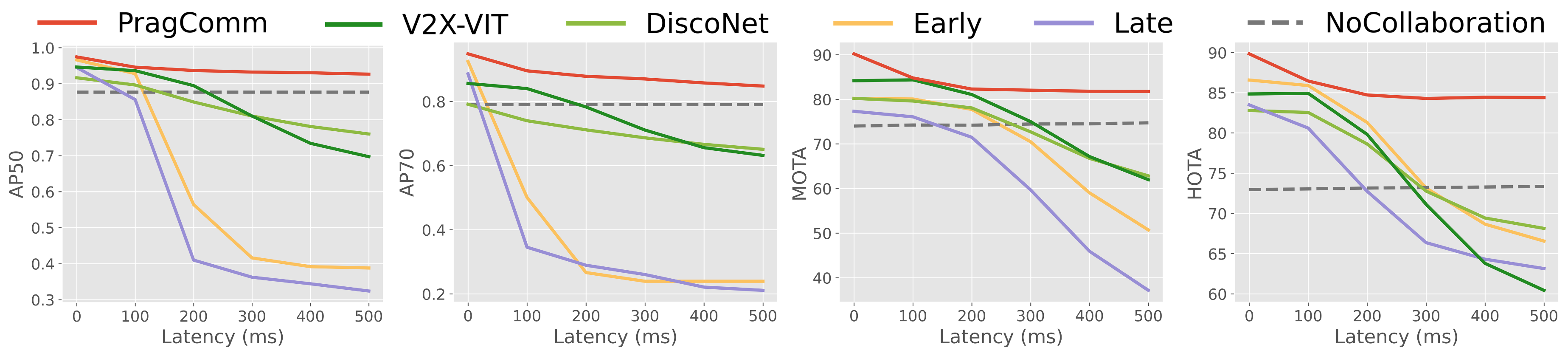}
  \vspace{-8mm}
  \caption{\texttt{PragComm} is robust to communication latency issue.}
  \vspace{-7mm}
  \label{Fig:Robust_latency}
\end{figure*}

\begin{figure}[!t]
\vspace{-1mm}
  \centering
  \includegraphics[width=1.0\linewidth]{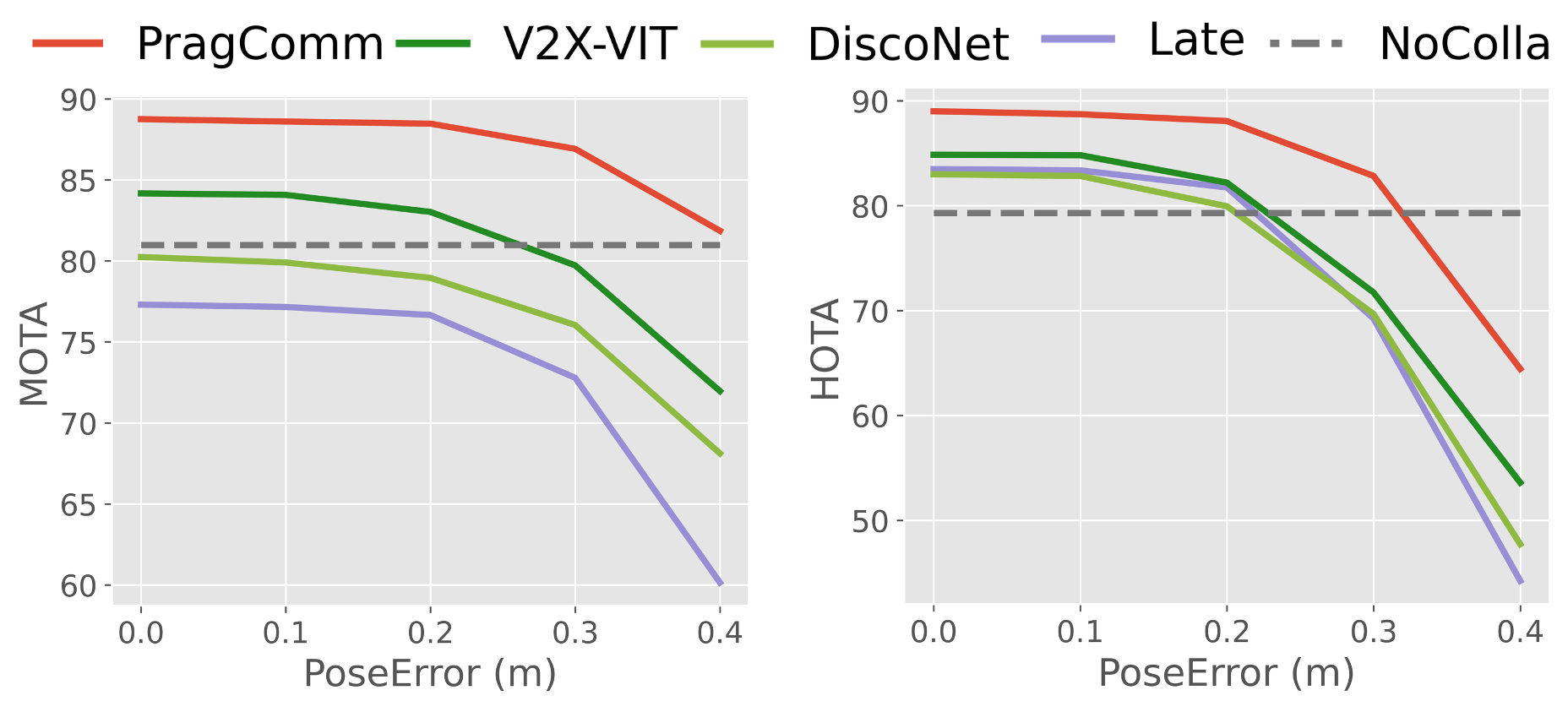}
  \vspace{-8mm}
  \caption{\texttt{PragComm} is robust to pose error issue.}
  \vspace{-6mm}
  \label{Fig:Robust_poseerror}
\end{figure}

\textbf{Effect of codebook configuration in channel compressor.} Fig.~\ref{Fig:ABL_Codebook} explore different codebook configurations: codebook size $n_L$ and code quantity $n_R$. We see that:
i) all configurations demonstrate a superior perception-communication trade-off, indicating that the codebook-based representation is robust to the configurations;
and ii) larger codebook sizes and quantities yield better performance, as they can approximate the original data with fewer quantization errors.

\textbf{Robustness to communication latency issue.} We validate the robustness against communication latency on both OPV2V. The latency setting follows SyncNet~\cite{LeiLatency:ECCV22}, varying from 0ms to 500ms.
Figs.~\ref{Fig:Robust_latency} show the detection performances as a function of pose error and latency, respectively. We see: i) while perception performance generally declines with increasing levels of latency, \texttt{PragComm} consistently outperforms baselines under all imperfect conditions; ii) \texttt{PragComm} consistently surpasses No Collaboration, whereas baselines fail when latency surpasses 300ms. The reason is that our message predictor can estimate the current state of a message, compensating for latency issues, while other baselines directly integrate the delayed latent messages, leading to misleading results.

\textbf{Robustness to pose error issue.}
We follow the pose error setting in V2VNet and V2X-ViT (Gaussian noise with a mean of 0m and a standard deviation of 0m-0.6m) to validate the robustness against realistic pose error. \texttt{PragComm} is more robust to the pose error than previous SOTAs. Fig.~\ref{Fig:Robust_poseerror} shows the detection performances as a function of pose error level. We see: i) overall the collaborative perception performance degrades with the increasing pose error, while \texttt{PragComm} outperforms previous SOTAs (V2X-ViT, DiscoNet) under all the pose errors. ii) \texttt{PragComm} keeps being superior to \texttt{No~Collaboration} while V2X-ViT fails when noise is over 0.3m. The reasons are: i) the codebook helps filter out noisy features, and ii) the message selection helps filter out noisy features; these two designs work together to mitigate noise localization distortion effects.

\begin{figure*}[!ht]
  \centering
  \begin{subfigure}{0.19\linewidth}
    \includegraphics[width=1.0\linewidth]{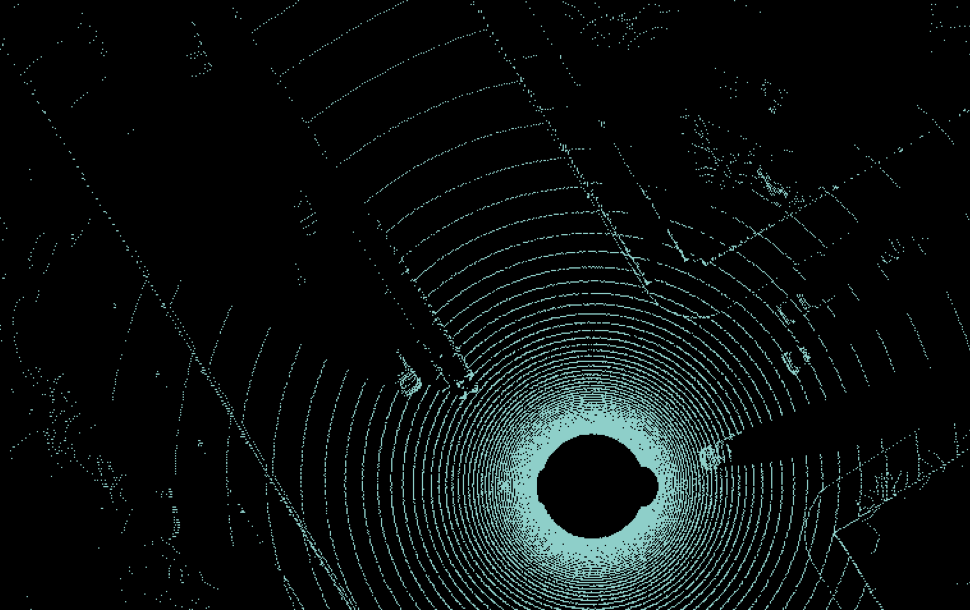}
    \vspace{-5mm}
    \caption{input $\mathcal{X}_i^{(t)}$}
  \end{subfigure}
  \begin{subfigure}{0.19\linewidth}
    \includegraphics[width=1.0\linewidth]{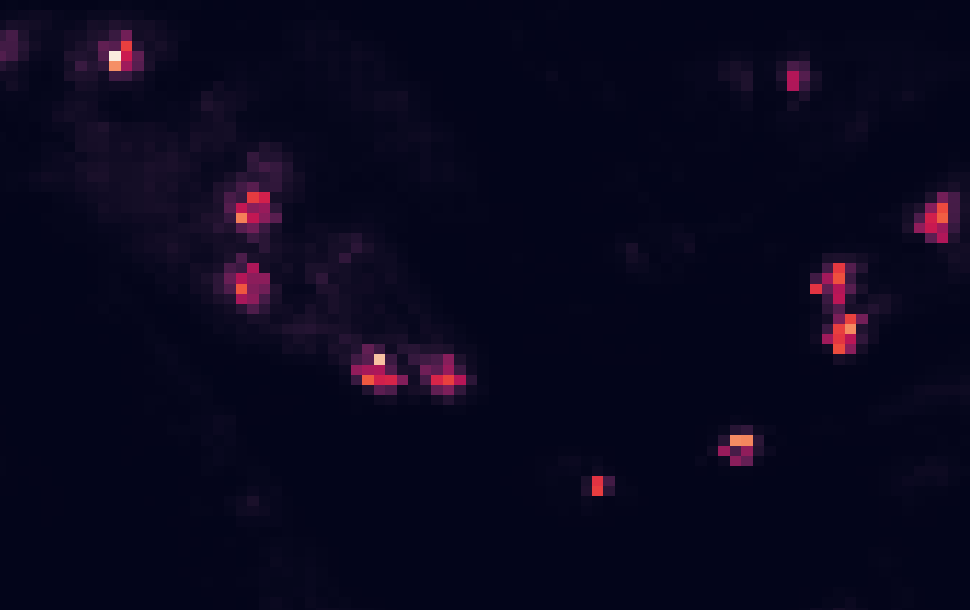}
    \vspace{-5mm}
    \caption{confidence map $\mathbf{C}_i^{(t)}$}
  \end{subfigure}
    \begin{subfigure}{0.19\linewidth}
    \includegraphics[width=1.0\linewidth]{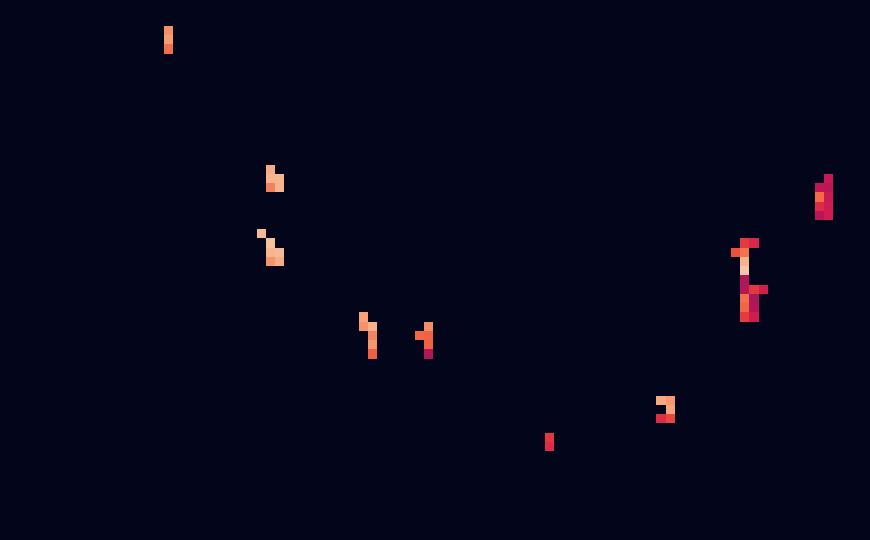}
    \vspace{-5mm}
    \caption{selected feature $\mathcal{Z}_{i\rightarrow j}^{(t)}$}
  \end{subfigure}
  \begin{subfigure}{0.19\linewidth}
    \includegraphics[width=1.0\linewidth]{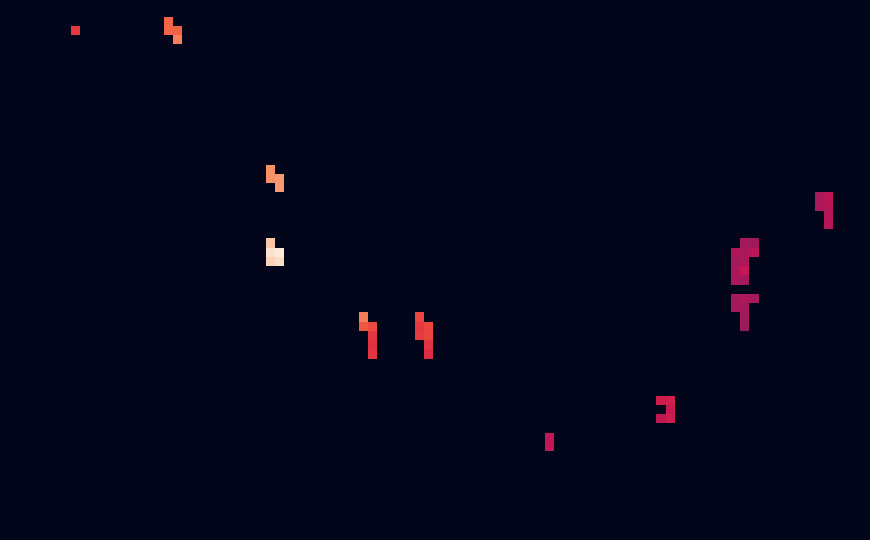}
    \vspace{-5mm}
    \caption{code feature $\overline{\mathcal{Z}}_{i\rightarrow j}^{(t)}$}
  \end{subfigure}
  \begin{subfigure}{0.19\linewidth}
    \includegraphics[width=1.0\linewidth]{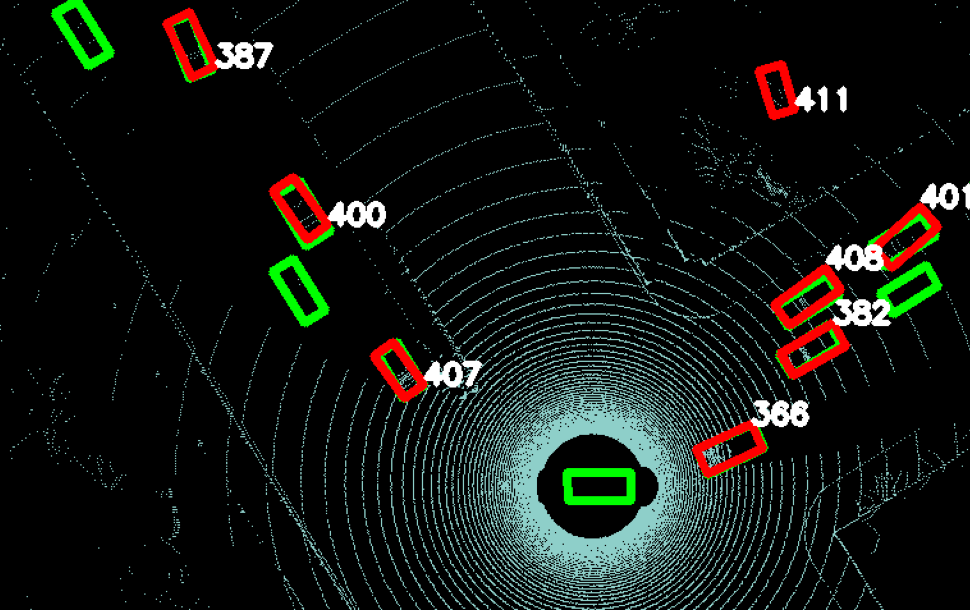}
    \vspace{-5mm}
    \caption{single detection $\mathcal{O}_i^{(t)}$}
  \end{subfigure}
    \vspace{1mm}

  \begin{subfigure}{0.19\linewidth}
    \includegraphics[width=1.0\linewidth]{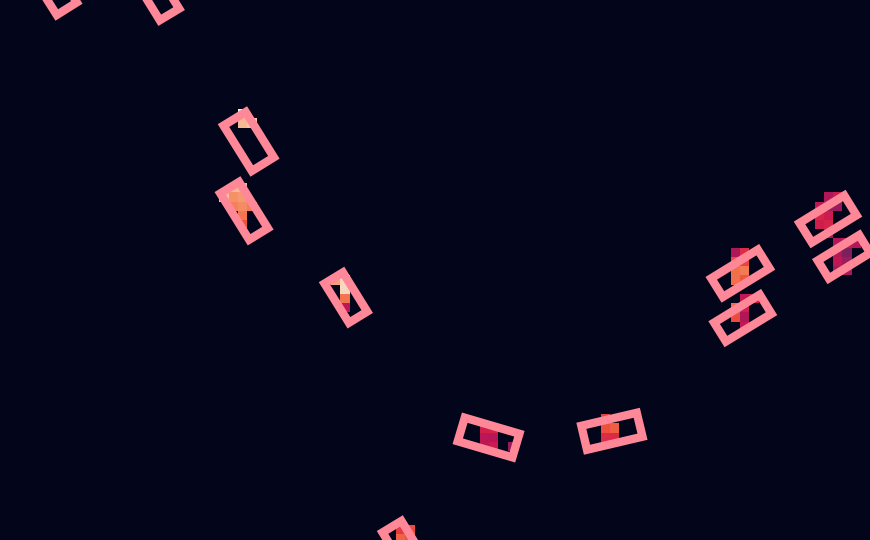}
    \vspace{-5mm}
    \caption{hist feature $\mathcal{H}_i^{(t-1)}$}
  \end{subfigure}
  \begin{subfigure}{0.19\linewidth}
    \includegraphics[width=1.0\linewidth]{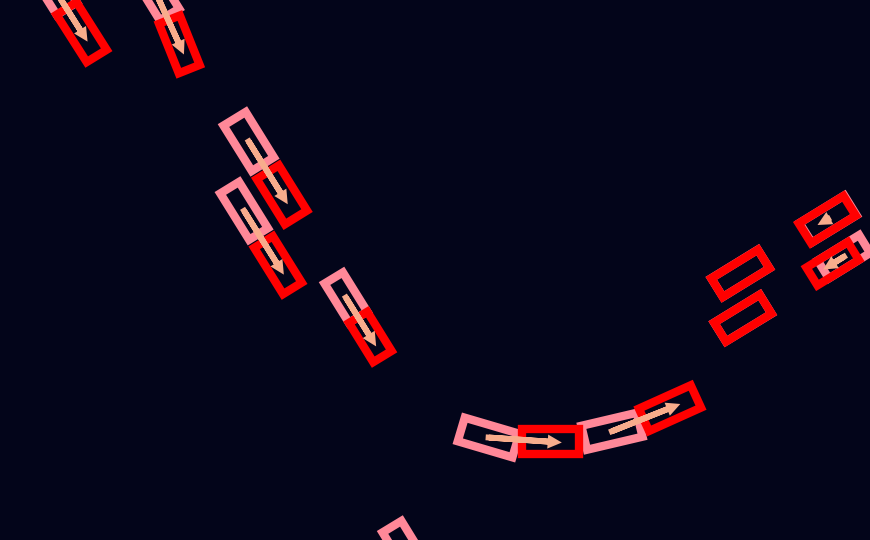}
    \vspace{-5mm}
    \caption{BEV flow $\mathcal{B}_i^{(t)}$}
  \end{subfigure}
    \begin{subfigure}{0.19\linewidth}
    \includegraphics[width=1.0\linewidth]{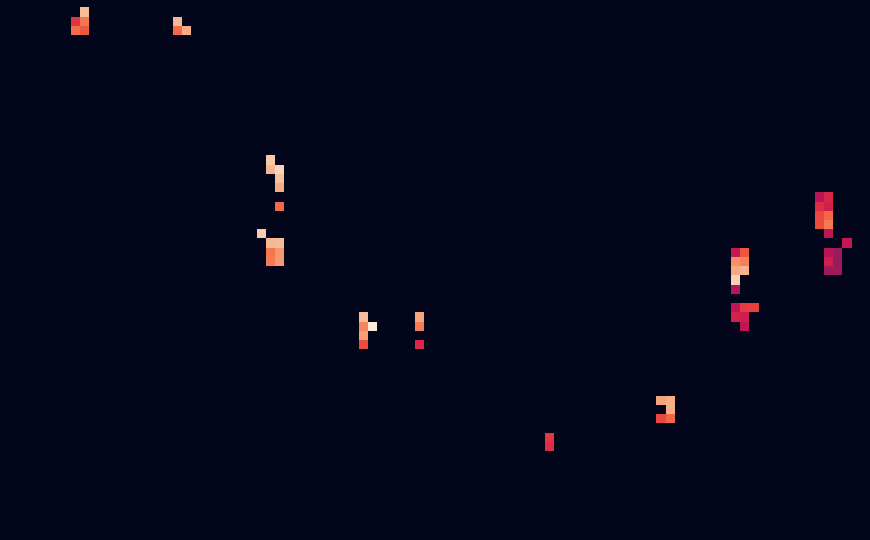}
    \vspace{-5mm}
    \caption{predicted feature $\overline{\mathcal{H}}_i^{(t)}$}
  \end{subfigure}
  \begin{subfigure}{0.19\linewidth}
    \includegraphics[width=1.0\linewidth]{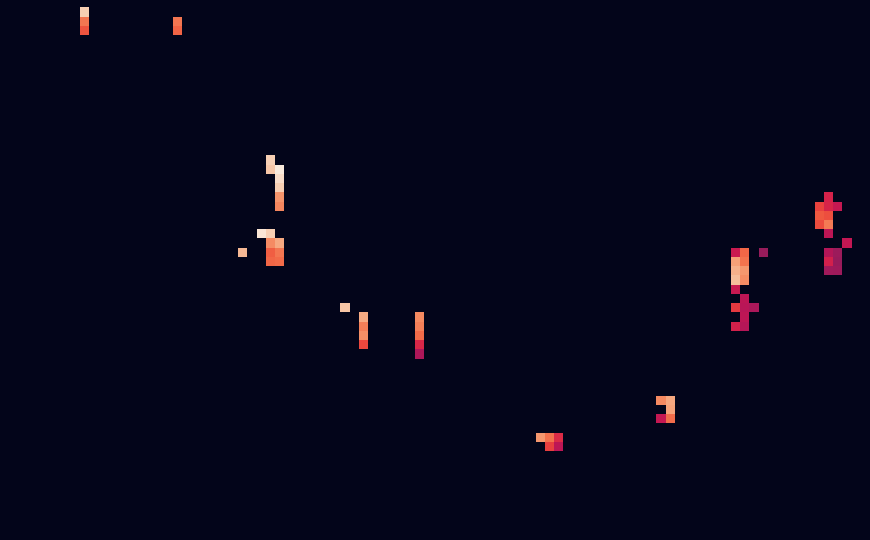}
    \vspace{-5mm}
    \caption{fused feature $\mathcal{H}_i^{(t)}$}
  \end{subfigure}
  \begin{subfigure}{0.19\linewidth}
    \includegraphics[width=1.0\linewidth]{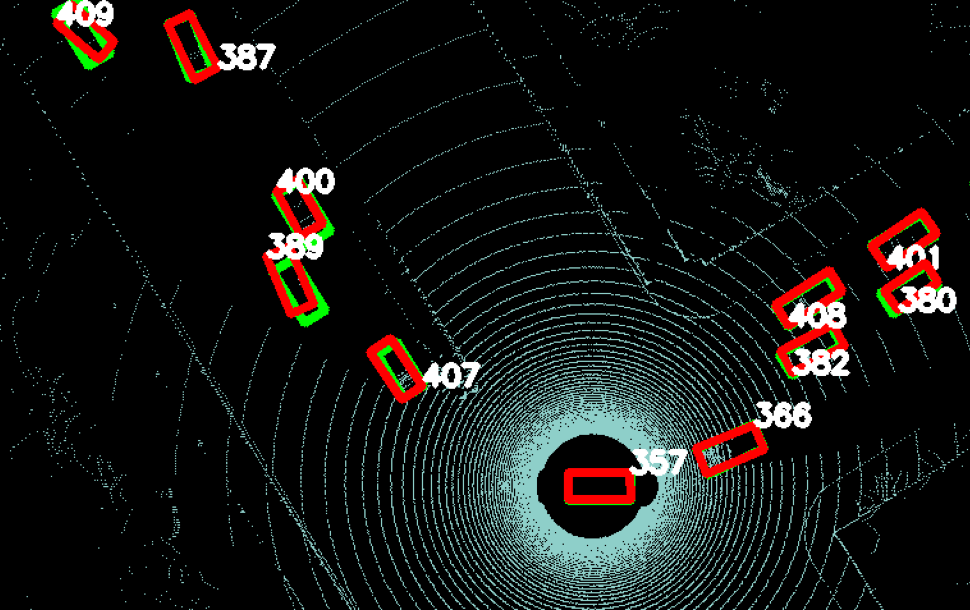}
    \vspace{-5mm}
    \caption{fused detection $\widehat{\mathcal{O}}_i^{(t)}$}
  \end{subfigure}
  
  \vspace{-3mm}
  \caption{Visualization of collaboration in \texttt{PragComm} on OPV2V dataset. \textcolor{green}{Green} and \textcolor{red}{red} boxes denote ground-truth and detection, respectively. The objects occluded in individual view can be detected by transmitting compact pragmatic messages.}
  \label{Fig:Collaboration}
  \vspace{-4mm}
\end{figure*}

\begin{figure*}[!ht]
  \centering
  \begin{subfigure}{0.24\linewidth}
    \includegraphics[width=1.0\linewidth]{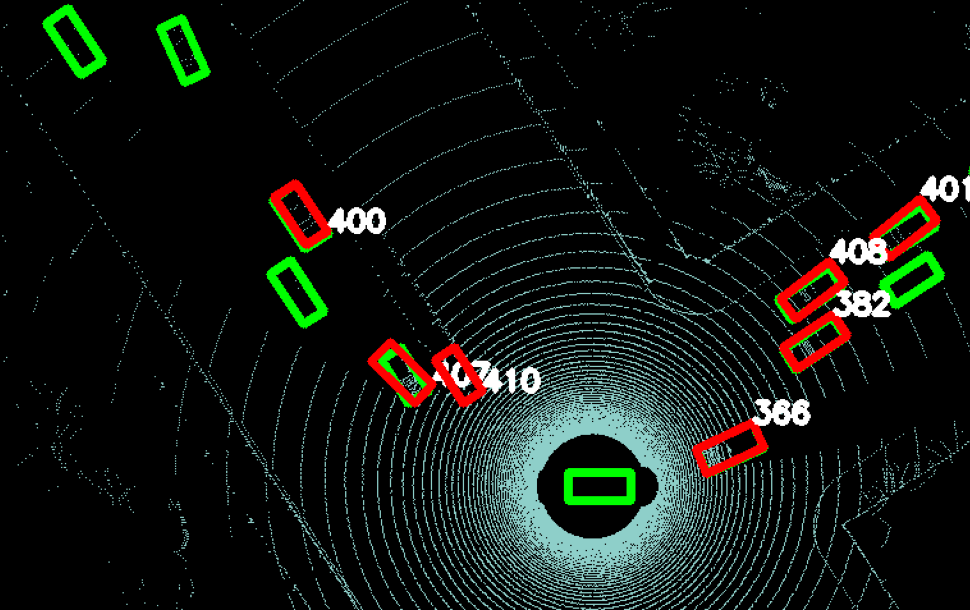}
  \end{subfigure}
  \begin{subfigure}{0.24\linewidth}
    \includegraphics[width=1.0\linewidth]{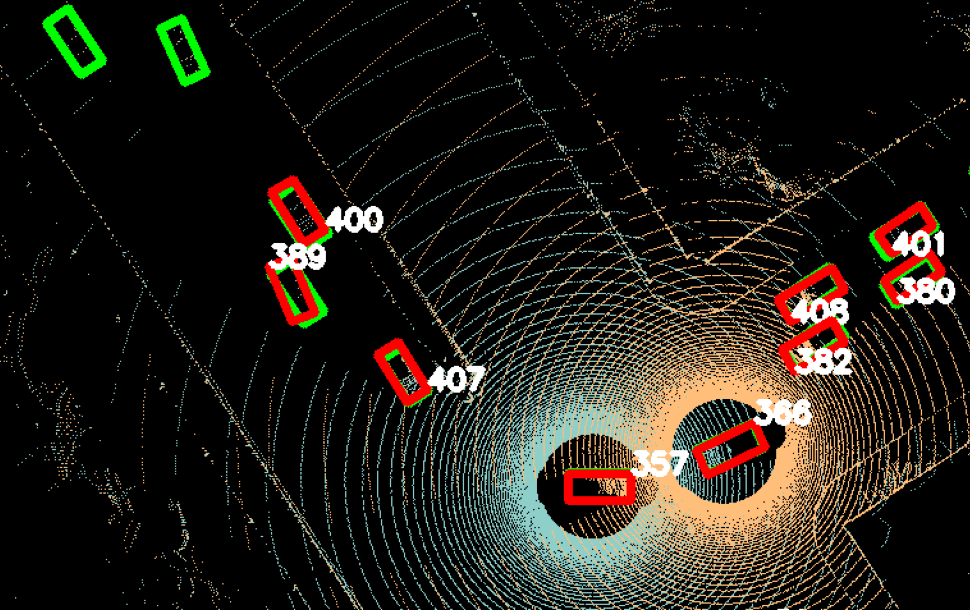}
  \end{subfigure}
    \begin{subfigure}{0.24\linewidth}
    \includegraphics[width=1.0\linewidth]{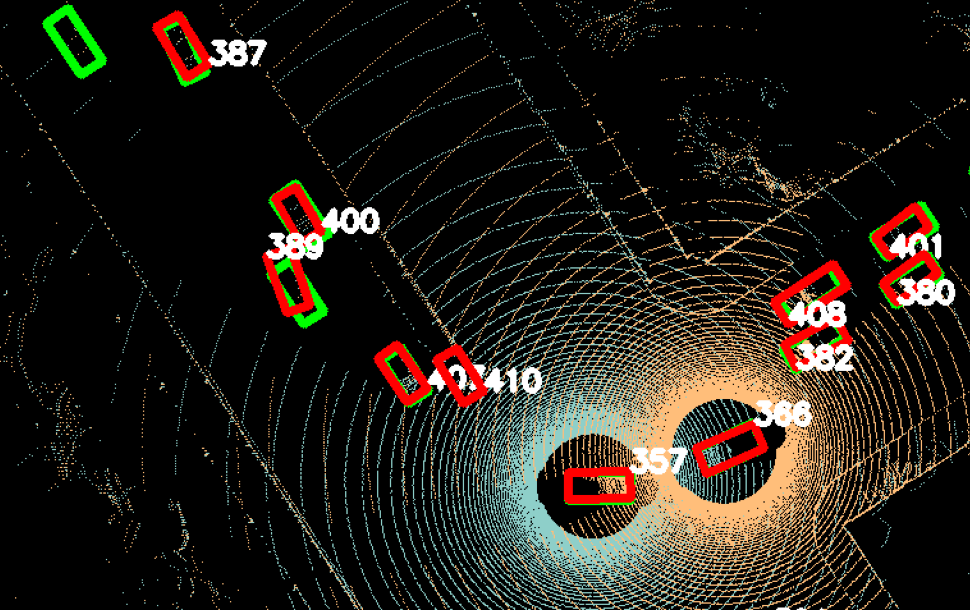}
  \end{subfigure}
  \begin{subfigure}{0.24\linewidth}
    \includegraphics[width=1.0\linewidth]{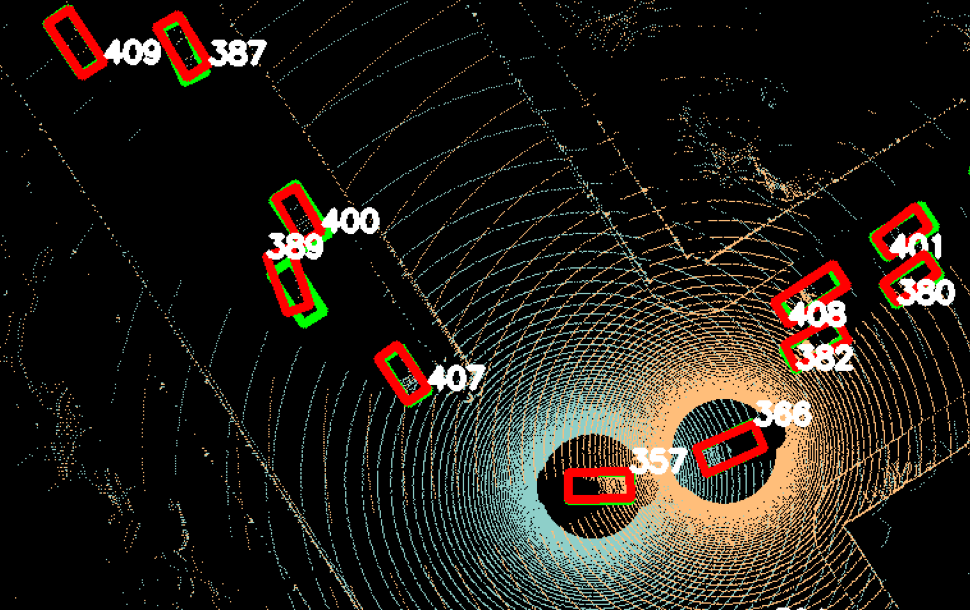}
  \end{subfigure}
    \vspace{1mm}

  \begin{subfigure}{0.24\linewidth}
    \includegraphics[width=1.0\linewidth]{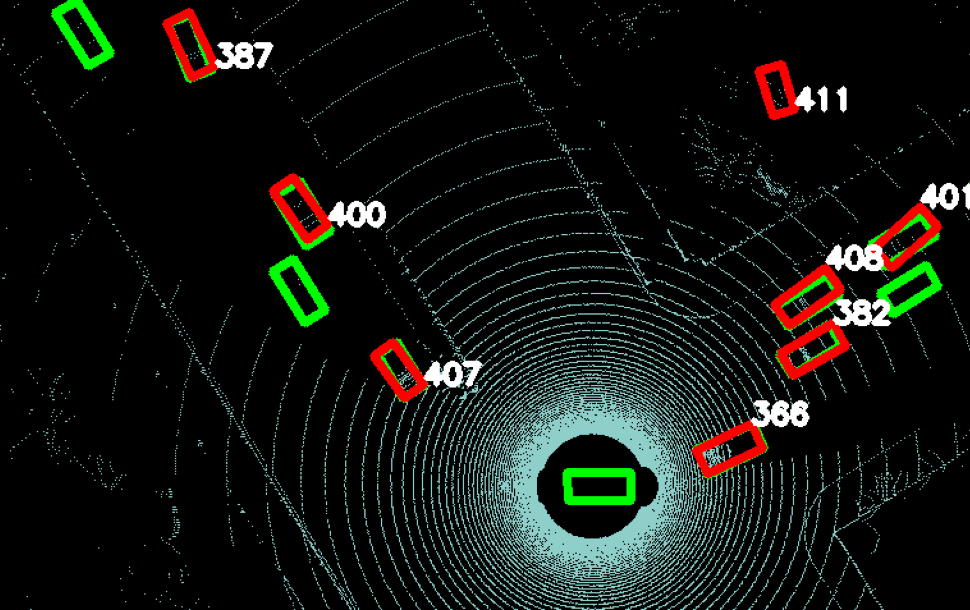}
    \vspace{-5mm}
    \caption{No Collaboration}
  \end{subfigure}
  \begin{subfigure}{0.24\linewidth}
    \includegraphics[width=1.0\linewidth]{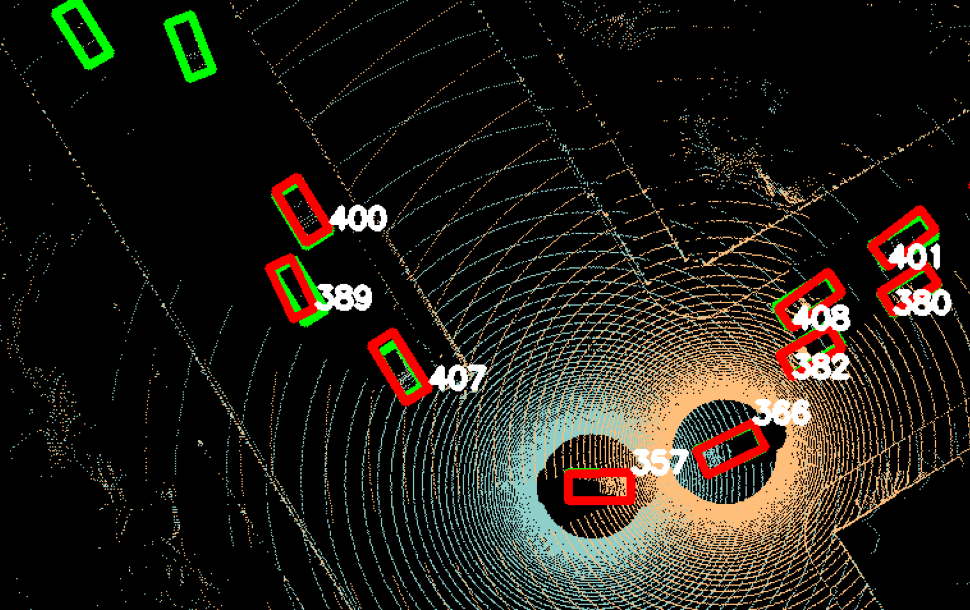}
    \vspace{-5mm}
    \caption{V2X-ViT}
  \end{subfigure}
    \begin{subfigure}{0.24\linewidth}
    \includegraphics[width=1.0\linewidth]{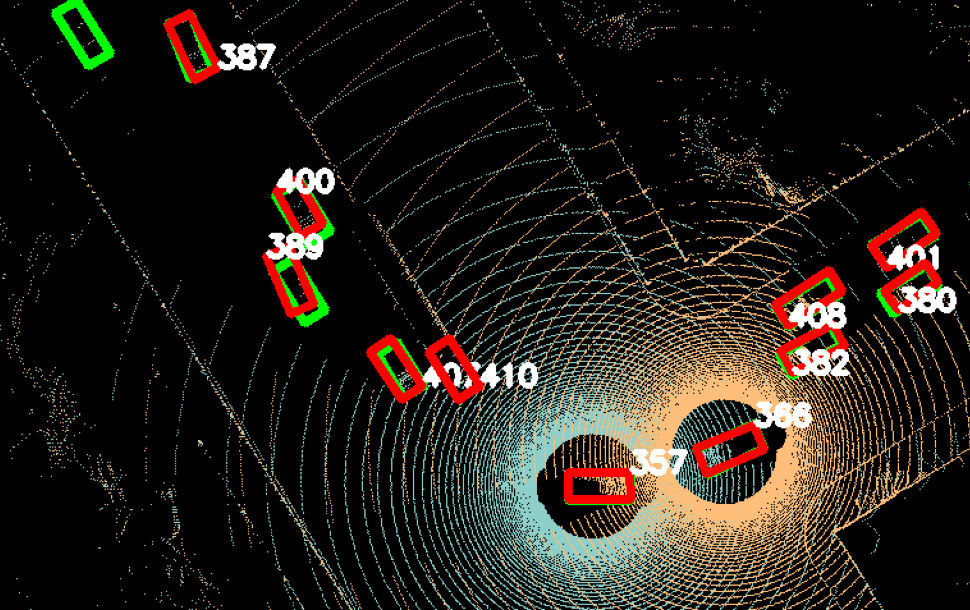}
    \vspace{-5mm}
    \caption{Where2comm}
  \end{subfigure}
  \begin{subfigure}{0.24\linewidth}
    \includegraphics[width=1.0\linewidth]{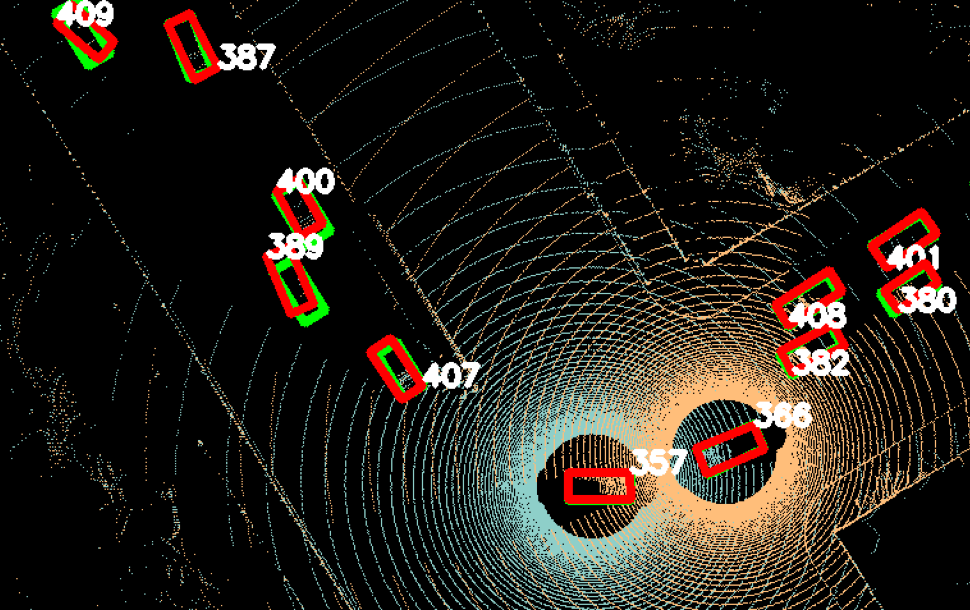}
    \vspace{-5mm}
    \caption{\texttt{PragComm}}
  \end{subfigure}
  
  \vspace{-3mm}
  \caption{\texttt{PragComm} qualitatively outperforms V2X-ViT and Where2comm for detection and tracking in OPV2V dataset. \textcolor{green}{Green} and \textcolor{red}{red} boxes denote ground-truth and detection, respectively. \textcolor[RGB]{173,216,230}{Blue} and \textcolor{orange}{orange} denote the point clouds collected from ego-agent and collaborator, respectively.}
  \label{Fig:opv2v_results}
  \vspace{-6mm}
\end{figure*}

\vspace{-2mm}
\subsection{Visualization}
\vspace{-1mm}
\textbf{Visualization of collaboration.} Fig.~\ref{Fig:Collaboration} illustrates how~\texttt{PragComm} benefits from communication-efficient collaboration. Through the proposed spatial, channel, and temporal compressors, perceptually critical spatial-temporal cues are selected and efficiently packed into pragmatic messages. This collaboration enables single agents to detect through occlusion.
The message compression process is illustrated in Fig.~\ref{Fig:Collaboration}, and the message decoding and fusion process is shown in Fig.~\ref{Fig:Collaboration} (f-i). Fig.~\ref{Fig:Collaboration} (a) displays the ego observation, which is subject to occlusion and long-range issues. Fig.~\ref{Fig:Collaboration} (b-c) depicts the spatial confidence map and sparse feature map output by the spatial compressor. The confidence map highlights spatial regions containing objects, which are retained after the spatial compression.
Fig.~\ref{Fig:Collaboration} (d) presents the discretized feature map generated by the channel compressor. The selected foreground areas share similar representations, efficiently represented with feature vectors from the compact codebook, and the corresponding code indices are packed in the final pragmatic messages.
Regarding the temporal compressor, when timestamps align with the sampling intervals, the packed messages are exchanged among agents. At these timestamps, agents directly look up the codebook and fuse the collaborative features. However, at other timestamps, agents do not receive fresh pragmatic messages. Instead, they collaborate with the available historical collaborative features, as depicted in Fig.~\ref{Fig:Collaboration} (f-i).
Fig.~\ref{Fig:Collaboration} (f) displays the historical collaborative feature from the previous timestamp. 
Fig.~\ref{Fig:Collaboration} (g) shows the estimated box flow, representing the object's movement pattern from the previous state to the current state. This information allows us to temporally compensate the historical feature to the current timestamp.
Fig.~\ref{Fig:Collaboration} (h) illustrates the compensated collaborative feature at the current timestamp.
The message fusion module then combines the ego feature and the compensated collaborative feature, resulting in the fused feature shown in Fig.~\ref{Fig:Collaboration} (i).
Fig.~\ref{Fig:Collaboration} (e) and (j) provide a comparison of the detection results before and after collaboration. 
We see that the proposed hyper-efficient message compressor generates sparse yet perceptually critical messages, effectively assisting agents in detecting occluded objects with greater efficiency.

\textbf{Visualization of detection and tracking results.} Fig.~\ref{Fig:opv2v_results} illustrates that~\texttt{PragComm} outperforms~\textit{No Collaboration},~\textit{V2XViT}, and~\texttt{Where2comm} in terms of more complete and accurate detection and tracking results while requiring lower communication cost.
The reasons behind this superiority are as follows:
i)~\texttt{PragComm} leverages temporally complementary historical pragmatic messages, enabling a more comprehensive perception compared to previous collaborative methods that only collaborate at the single current timestamp;
ii)~\texttt{PragComm} enhances communication efficiency by sampling informative spatial regions and essential timestamps and employing an efficient codebook-based feature indice representation. In contrast,~\texttt{Where2comm} focuses only on spatial sampling, and~\texttt{V2XViT} indistinctively shares complete spatial regions at all timestamps with costly feature vector representation;
and iii) collaborative perception enables complementary information sharing among agents, allowing a more comprehensive perception of regions that were previously obscured in single-agent views but are visible in the views of collaborators.

\vspace{-2mm}
\section{Conclusion}
\label{sec:conclusion}


In this paper, we propose a novel pragmatic communication strategy that specifies pragmatic messages for each agent’s specific perception task demand. These pragmatic messages selectively discard a large portion of task-irrelevant data. The amount of communication volume breaks Shannon’s limits, leading to enhanced communication efficiency while maintaining perception utility. 
Based on this novel strategy, we propose~\texttt{PragComm}, a novel pragmatic collaborative perception system, which uses pragmatic messages to enhance the detection and tracking capabilities of multiple collaborative agents. 
We validate the effectiveness of \texttt{PragComm} with extensive experiments on various real-world and simulated datasets. Experimental results show that \texttt{PragComm} achieves superior perception-communication trade-off across varying bandwidth conditions and multiple perception tasks.



\vspace{-3mm}
\section*{Acknowledgments}
\vspace{-1mm}
This research is supported by the National Key R\&D Program of China under Grant 2021ZD0112801, NSFC under Grant 62171276 and the Science and Technology Commission of Shanghai Municipal under Grant 21511100900 and 22DZ2229005.

\vspace{-3mm}
\bibliographystyle{IEEEtran}
\bibliography{main}

\vspace{-35mm}
\begin{IEEEbiography}[{\includegraphics[width=1in,height=1.25in,clip,keepaspectratio]{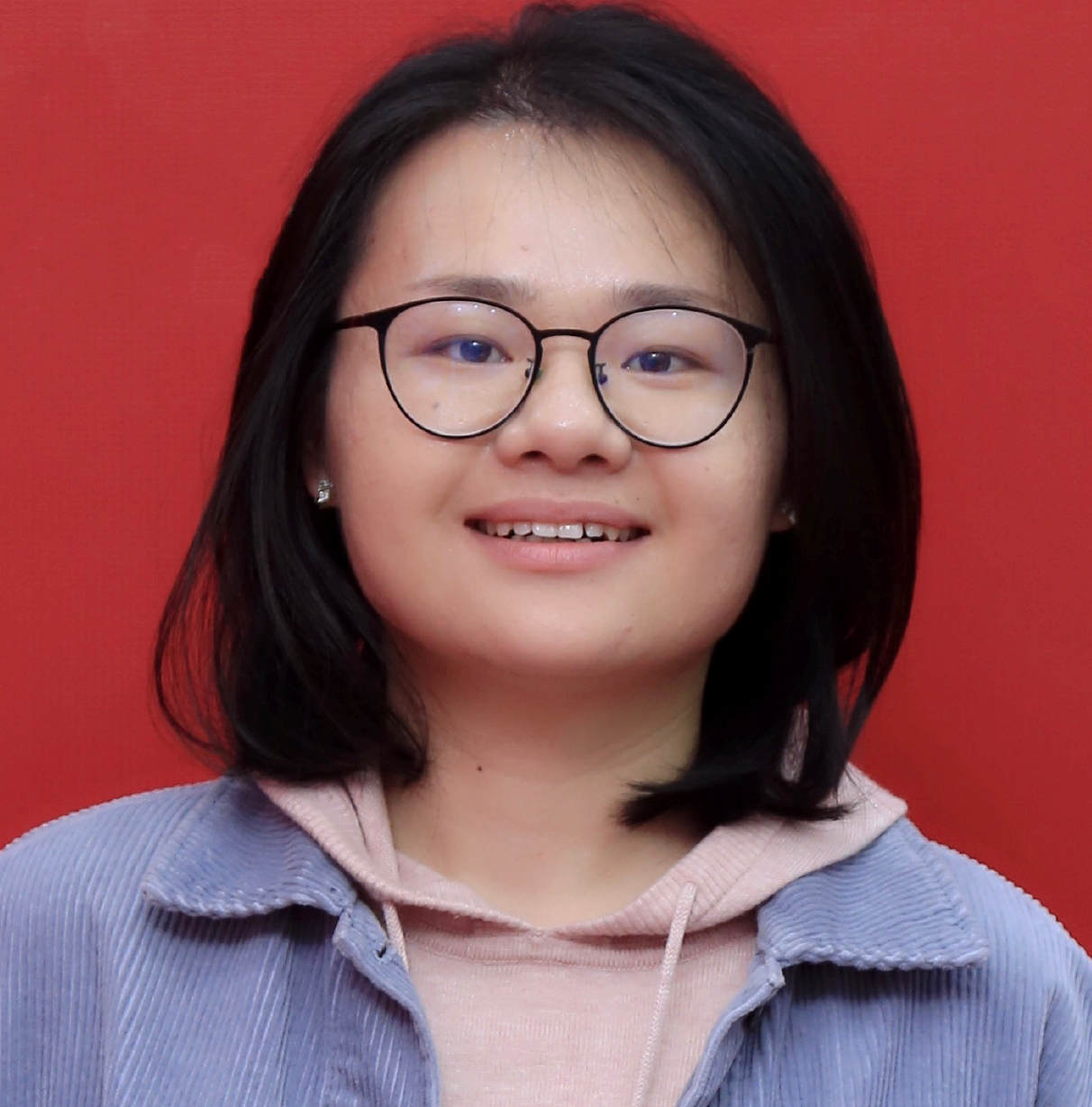}}]{Yue Hu} is working toward the Ph.D. degree at Cooperative Medianet Innovation Center at Shanghai Jiao Tong University since 2021. She received the M.S. degree and B.E. degree in information engineering from Shanghai Jiao Tong University, Shanghai, China, in 2020 and 2017. Her research interests include multi-agent collaboration, communication efficiency, and 3D vision.
\end{IEEEbiography}
\vspace{-40mm}

\begin{IEEEbiography}[{\includegraphics[width=1in,height=1.25in,clip,keepaspectratio]{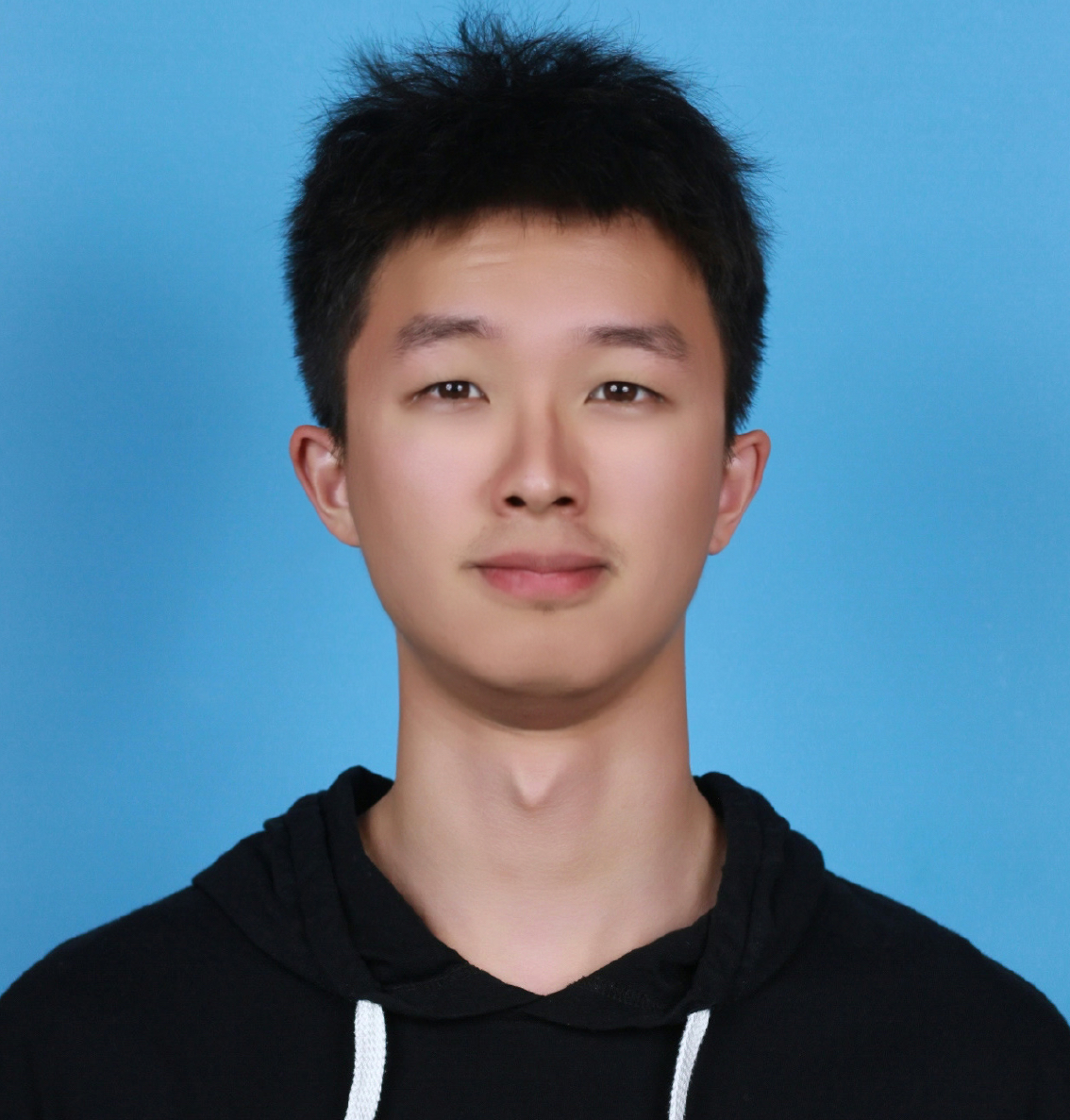}}]{Xianghe Pang} received the B.E. degree in Electronic and Information Engineering from University of Electronic Science and Technology of China, Sichuan, China, in 2022. He is working toward the Ph.D. degree at Cooperative Medianet Innovation Center at Shanghai Jiao Tong University since 2022. His research interests include multi-agent collaboration and communication efficiency.
\end{IEEEbiography}

\vspace{-35mm}

\begin{IEEEbiography}[{\includegraphics[width=1in,height=1.25in,clip,keepaspectratio]{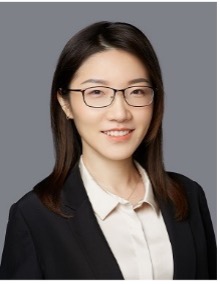}}]{Xiaoqi Qin} received her B.S., M.S., and Ph.D. degrees from Electrical and Computer Engineering with Virginia Tech. She is currently an Associate Professor of School of Information and Communication Engineering with Beijing University of Posts and Telecommunication (BUPT). She has received the Best Paper Awards at IEEE GLOBECOM’23 and WCSP’23. Her research mainly focuses on task-oriented machine-type communications and networked intelligence. She was a recipient of first Prize of Science and Tech. Progress Award by Chongqing Municipal People's Government, and first Prize of Tech. Invention Award by China Institute of Communications.
\end{IEEEbiography}

\begin{IEEEbiography}
[{\includegraphics[width=1in,height=1.25in, clip,keepaspectratio]{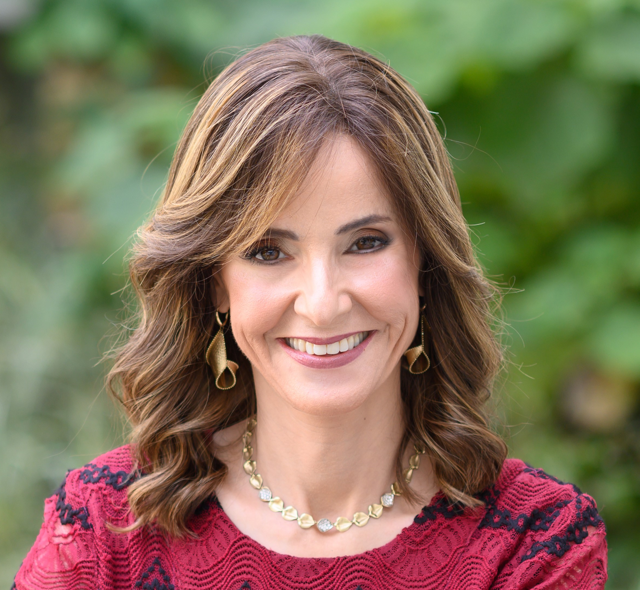}}]{Yonina C. Eldar} (Fellow, IEEE) received the B.Sc. degree in Physics in 1995 and the B.Sc. degree in Electrical Engineering in 1996 both from Tel-Aviv University (TAU), Tel-Aviv, Israel, and the Ph.D. degree in Electrical Engineering and Computer Science in 2002 from the Massachusetts Institute of Technology (MIT), Cambridge. She is currently a Professor in the Department of Mathematics and Computer Science, Weizmann Institute of Science, Rehovot, Israel where she holds the Dorothy and Patrick Gorman Professorial Chair and heads the Center for Biomedical Engineering. Dr. Eldar has received many awards for excellence in research and teaching, including the  IEEE Signal Processing Society Technical Achievement Award (2013), the IEEE/AESS Fred Nathanson Memorial Radar Award (2014), and the IEEE Kiyo Tomiyasu Award (2016). She is the Editor in Chief of Foundations and Trends in Signal Processing, a member of the IEEE Sensor Array and Multichannel Technical Committee and serves on several other IEEE committees. She is a member of the Israel Academy of Sciences and Humanities (elected 2017) and of the Academia Europaea (elected 2023), an IEEE Fellow, a EURASIP Fellow, a Fellow of the Asia-Pacific Artificial Intelligence Association, and a Fellow of the 8400 Health Network. Her research interests are in the broad areas of statistical signal processing, sampling theory and compressed sensing, learning and optimization methods, and their applications to biology, medical imaging and optics.
\end{IEEEbiography}

\vspace{-20mm}
\begin{IEEEbiography}[{\includegraphics[width=1in,height=1.25in,clip,keepaspectratio]{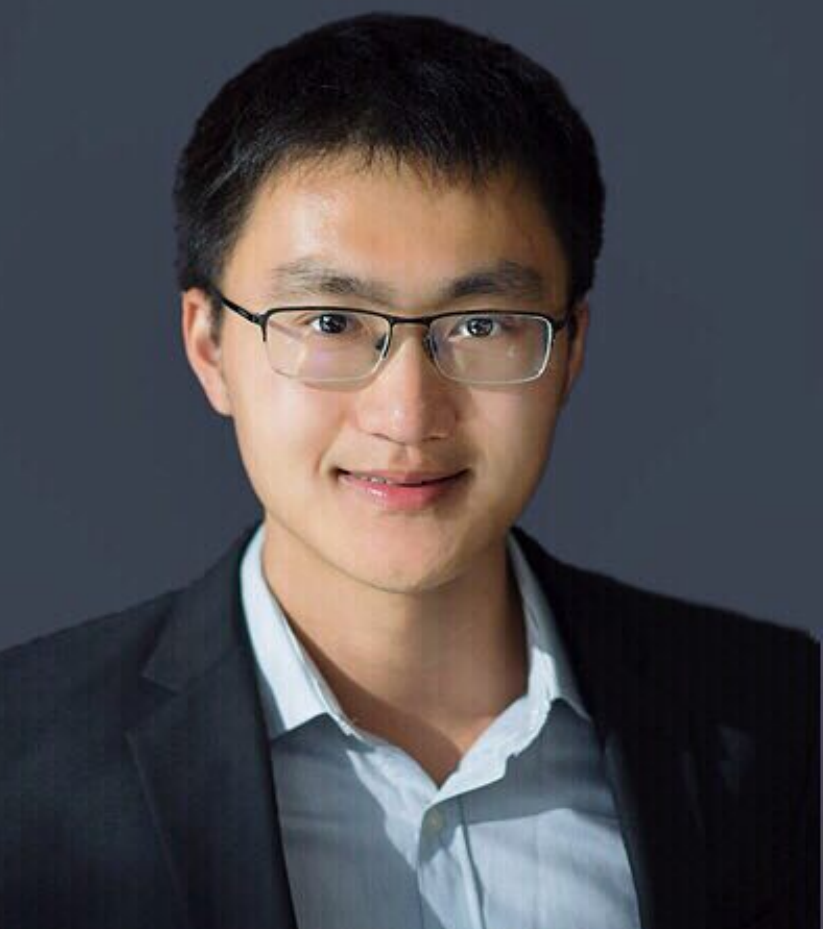}}]{Siheng Chen} is a tenure-track associate professor of Shanghai Jiao Tong University. He was a research scientist at Mitsubishi Electric Research Laboratories (MERL), and an autonomy engineer at Uber Advanced Technologies Group (ATG), working on self-driving cars. Dr. Chen received his doctorate from Carnegie Mellon University in 2016. Dr. Chen's work on sampling theory of graph data received the 2018 IEEE Signal Processing Society Young Author Best Paper Award. He contributed to the project of scene-aware interaction, winning MERL President's Award. His research interests include autonomous driving and collective intelligence.
\end{IEEEbiography}

\vspace{-20mm}
\begin{IEEEbiography}[{\includegraphics[width=1in,height=1.25in,clip,keepaspectratio]{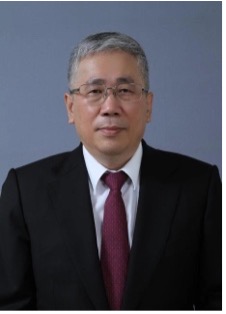}}]{Ping Zhang} is currently a professor of School of Information and Communication Engineering at Beijing University of Posts and Telecommunications, the director of State Key Laboratory of Networking and Switching Technology. He served as Chief Scientist of National Basic Research Program (973 Program), an expert in Information Technology Division of National High-tech R\&D program (863 Program), a member of IMT-2020 (5G) Experts Panel, a member of Experts Panel for China’s 6G development, and a member of Consultant Committee on International Cooperation of National Natural Science Foundation of China. His research interests mainly focus on wireless communication. He is an Academician of the Chinese Academy of Engineering (CAE).
\end{IEEEbiography}

\vspace{-20mm}
\begin{IEEEbiography}[{\includegraphics[width=1in,height=1.25in,clip,keepaspectratio]{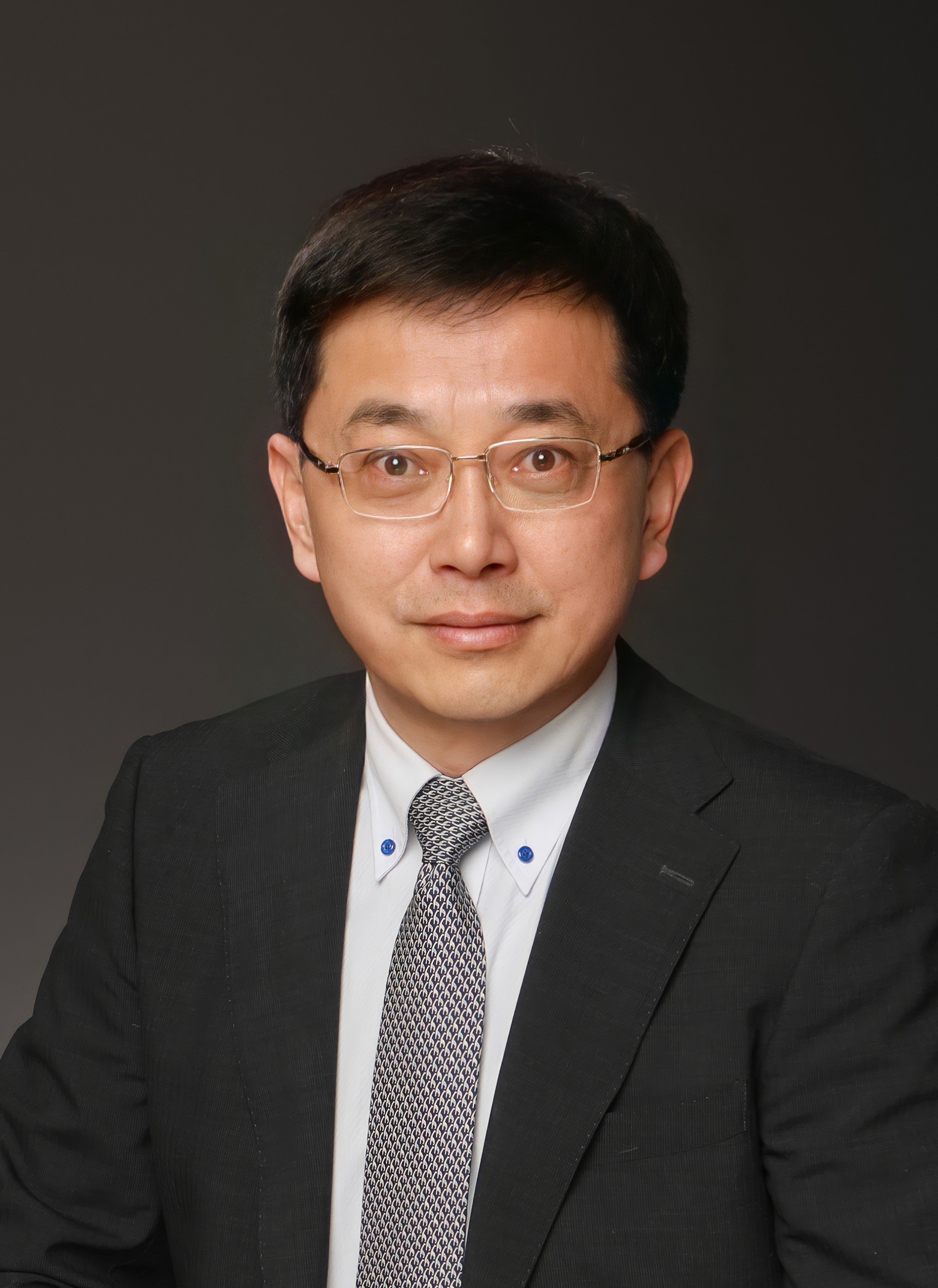}}]{Wenjun Zhang} (Fellow, IEEE) received the B.S., M.S., and Ph.D. degrees in electronic engineering from Shanghai Jiao Tong University, Shanghai, China, in 1984, 1987, and 1989, respectively. He is now a Full Professor professor at Shanghai Jiao Tong University. His main research interests include digital video coding and transmission, multimedia semantic processing, and intelligent video surveillance. He is a Chief Scientist with the Chinese National Engineering Research Centre of Digital Television (NERC-DTV), an industry/government consortium in DTV technology research and standardization, and the Chair of the Future of Broadcast Television Initiative Technical Committee. As the national HDTV TEEG project leader, he developed the first Chinese HDTV prototype system in 1998. He was one of the main contributors to the Chinese Digital Television Terrestrial Broadcasting Standard issued in 2006 and has been leading team in designing the next generation of broadcast television system in China since 2011.
\end{IEEEbiography}
\vspace{-10mm}



\end{document}